\documentclass[10pt,twocolumn,letterpaper]{article}

\usepackage{iccv}
\usepackage{times}
\usepackage{epsfig}
\usepackage{graphicx}
\usepackage{amsmath}
\usepackage{amssymb}
\usepackage{subfigure}
\usepackage{rotating}
\usepackage{multirow}

\usepackage[pagebackref=true,breaklinks=true,letterpaper=true,colorlinks,bookmarks=false]{hyperref}

\iccvfinalcopy 

\def\iccvPaperID{5099} 

\newcommand{\vect}[1]{\boldsymbol{\mathbf{#1}}}
\ificcvfinal\fi
\begin{document}

\title{A Differential Volumetric Approach to Multi-View Photometric Stereo}

\author{Fotios Logothetis$^{1,2}$, \quad Roberto Mecca$^2$, \quad Roberto Cipolla$^1$\\
{\tt\small fl302@cam.ac.uk, roberto.mecca@crl.toshiba.co.uk, rc10001@cam.ac.uk}\\
$^1$Department of Engineering, University of Cambridge, United Kingdom\\
$^2$Toshiba Research, Cambridge, United Kingdom\\
}

\maketitle

\begin{abstract}
\vspace{-0.25cm}
Highly accurate 3D volumetric reconstruction is still an open research topic where the main difficulty is usually related to merging some rough estimations  with high frequency details. One of the most promising methods is the fusion between multi-view stereo and photometric stereo images. Beside the intrinsic difficulties that multi-view stereo and photometric stereo in order to work reliably, supplementary problems arise when considered together. 

In this work, we present a volumetric approach to the multi-view photometric stereo problem. The key point of our method is the signed distance field parameterisation and its relation to the surface normal. This is exploited in order to obtain a linear partial differential equation which is solved in a variational framework, that combines multiple images from multiple points of view in a single system. In addition, the volumetric approach is naturally implemented on an octree, which allows for fast ray-tracing that reliably alleviates occlusions and cast shadows.

Our approach is evaluated on synthetic and real data-sets and achieves state-of-the-art results.




\end{abstract}


	\vspace{-0.5cm}
\section{Introduction}
\label{sec:introduction}

Recovering the 3D geometry of an object is still a quite open challenge in computer vision as most of the techniques provide good results in specific frameworks only. In particular, two well-known approaches namely multi-view (MVS) and photometric stereo (PS) have been developed to produce great results considering some key complementary assumptions. Indeed, while 
MVS is assumed to provide rough 3D volumetric reconstructions of textured objects, PS 
is supposed to retrieve highly detailed surfaces from a single view. High quality volumetric reconstruction of objects achieved by refining coarse multi-view reconstruction \cite{Harltey2006,seitz2006comparison} with shading information \cite{Horn1,Woodham1980,Shi2018} is a classical way \cite{Blake1985} of combining complementary information. 

Multi-View Photometric Stereo (MVPS) approaches have been developed so as to overcome constraints coming from both sides, in order to deal with: specular highlights \cite{Jin2003,Ackermann2014}, dynamic scenes \cite{Vlasic2009}, 
visibility and occlusions \cite{Delaunoy2011}
and mapping of the PS views onto the 
volume \cite{Sabzevari2012,Park2017}. 

\begin{figure}[t]
 	\centering   
\includegraphics[clip,width=0.49\textwidth,trim={0.5cm 1.0cm 0.1cm  0.5cm}]{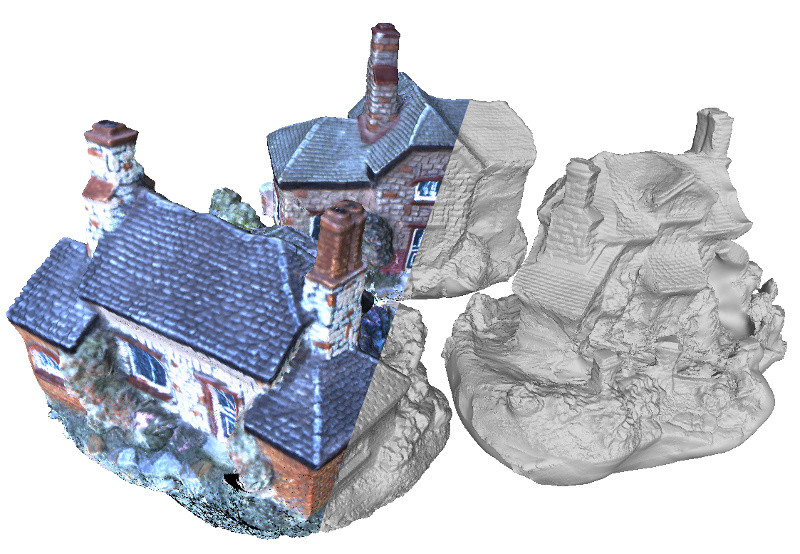}
\vspace{0.05cm}
\includegraphics[width=0.155\textwidth]{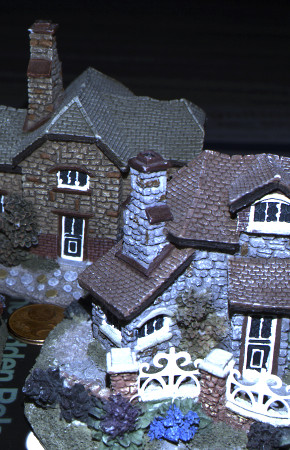}
\hspace{0.025cm}
\includegraphics[width=0.155\textwidth]{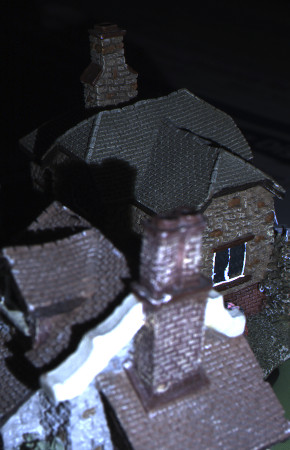}
\hspace{0.025cm}
\includegraphics[width=0.155\textwidth]{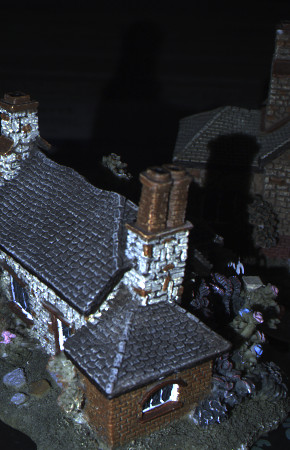}
  \caption{Top: 3D reconstruction of a three (toy) house scene with and without albedo, respectively left and right side. Bottom row: sample images. The scene scale is indicated by the coin on left bottom image. Our approach is capable of dealing with the near-field effects (point light sources,  out-of-focus shapes (\textit{middle})), occlusions and cast shadows.
  }
   \label{fig:real_data_houses_intro}
\end{figure}

Since implicit parameterisation of volumes has been developed using level-set approaches \cite{MalladiSV95,osher2006level}, recent advances in parameterising volumes with signed distance functions (SDFs) \cite{Zollhofer15,NieBner2013} have made the multi-view approach prone to be merged with differential formulation of irradiance equation providing shading information \cite{Maier2017}. On the other hand, recent photometric stereo approaches have moved towards more realistic assumptions considering point light sources \cite{Mecca2014near,queau2018led} that make the acquisition process easier by using LEDs in a calibrated setting.

In this work, our aim is to propose a new MVPS approach capable of dealing with scene having complex geometry when acquiring images in the near-field. This increases the difficulty of the reconstruction problem due to the severe visibility issues given by occlusions, cast shadows, out-of-focus regions, etc. (see Figure \ref{fig:real_data_houses_intro}).  
\vspace{-0.25cm}
\paragraph{\textbf{Contribution}} To do so, our novel approach is based on
\begin{itemize}

\item A differential parameterisation of the volume, based on the signed distance function that
allows irradiance equation ratios to deal with near-field Photometric Stereo modelling.

\item A variational optimisation that fuses information from multiple viewpoints into a single
system. 

\item An octree implementation allowing for quick raytracing so as highly accurate volumetric reconstructions can be obtained in scenes with multiple  objects.
\end{itemize} 

 	\section{Related Works}
\label{sec:relatedworks}

Reconstructing accurate 3D geometry of a volume has been a challenging area in computer vision. Most of the research trying to solve this problem has been developed by merging multi-view methods for coarse reconstruction \cite{Harltey2006}, with techniques based on shading information for providing high frequency details of the surface \cite{Moses2009,Weber2002,WuLiu2011,nehab2005efficiently,Beljan2012,GuoTOG2017} rather than based on a topological evolution of the surface \cite{Jin2003}. However, regarding the refinement, several methods take inspiration from Shape from Shading \cite{Horn1} to extract 3D geometry from a single image (MVSfS) and consider shape refinement resulting from single shading cues \cite{Wu2014shading,Wu2011Shading,Barron2015}. With the aim to improve the quality of the details and make the reconstruction more robust to outliers, multiple shading images from a single view point are considered. A number of MVPS approaches have been presented \cite{Hernandez2008Multiview,Park2017,Zhou2013multi}.

Merging shading information with multi-view images becomes a more complicated problem when considering specular surfaces. Drastic changes in both shading under different lighting and the viewing point modify the appearance of the 3D geometry so that specific approaches have been developed to deal with irradiance equations with not negligible specular component. Jin \emph{et al.} \cite{Jin2003} exploit a rank constraint on the radiance tensor field of the surface in space with the aim to fit the Ward reflectance model \cite{Ward92}. Other approaches instead reconstructed an unknown object by using a radiance basis inferred from reference objects \cite{Treuille2004,Ackermann2014}. Zhou \emph{et al.} \cite{Zhou2013multi} developed a camera and a handheld moving light system for firstly capturing sparse 3D points and then refining the depth along iso-depth contours \cite{Alldrin2007Towards}. A similar handheld system has been developed by Higo \emph{et al.} \cite{Higo2009} where multi-view images were acquired under varying illumination by a handled camera with a single movable LED point light source for reconstructing a static scene. 

In order to make the MVPS solvable, additional assumptions have been considered. In particular, with the aim to compute the camera positions so as to map accurately the photometric stereo views, the relative motion of the camera and the object can be constrained. Hernandez \emph{et al.} \cite{Hernandez2008Multiview} captured multi-view images for a moving object under varying illuminations by combining shading and silhouettes assuming circular motion in order to compute the visual hull. Zhang \emph{et al.} \cite{Zhang2003} generalised optical flow, photometric stereo, multiview-stereo and structure from motion techniques assuming rigid motion of the object under orthographic viewing geometry and directional lighting. Furthermore, shadows and occlusions are not considered.

When photometric stereo (as well as SfS) has to be integrated with multi-view techniques, the problem of finding the correspondence of pixels with shading information onto the 3D surface is crucial. Geometric distortions produced by changes in pose have to be combined with varying illumination. One way to do so is by region tracking considering brightness variations using optical flow \cite{Haussecker2001}, parametric models of geometry and illumination \cite{Hager1998}, or outlier rejection \cite{Jin2001}. Okatani and Deguchi \cite{Okatani2012} proposed a photometric method for estimating the second derivatives of the surface shape of an object when only inaccurate knowledge of the surface reflectance and illumination is given by assuming represented in a probabilistic fashion. 

Other approaches instead align the shading images with the coarse 3D in order to map the photometric stereo data onto the 3D shape \cite{Lim2005,Joshi2007}. Delaunoy and Prados \cite{Delaunoy2011} use a gradient flow approach whereas Sabzevari \emph{et. al} \cite{Sabzevari2012} firstly computes a 3D mesh with structure from motion with a low percentage of missing point and then the mesh is reprojected onto a plan using a mapping scheme \cite{Liu2008}. Recently, Park \emph{et al.} \cite{Park2017} proposed a refinement method by computing an optical displacement map in the same 2D planar domain of the photometric stereo images. So, they transformed the coarse 3D mesh into a parameterised 2D space using a parameterization technique that reduces distortions \cite{Sheffer2006}.  

In this work, with the aim to avoid the mapping procedure, we present a differential approach for MVPS. Being inspired by the signed distance function parameterisation used by Maier \emph{et al.} \cite{Maier2017} for the MVSfS problem, we derive a volumetric parameterisation which handles the differential irradiance equation ratio presented in
\cite{Mecca2014near} for near-field photometric stereo. The problem is posed in a 3D domain which in practice is implemented in a octree, which allows for fast ray-tracing. This accelerates the computation of cast shadows (that are similarly conceived as in \cite{logothetis2017semi}) and occlusions from different views and makes it possible to generate sub-milimeter precision models for scenes occupying a volume of several liters.



    \section{Signed Distance Function Parameterisation}
\label{sec:mathematical}

With the aim to provide suitable mathematical characterisation of a collection of solid objects, we consider the implicit surface parameterisation in terms of the SDF $d(\vect{x}), \vect{x} \in \mathbb{R}^3$. This parameterisation turns out to be suitable for our aim due to its practical way of describing the outgoing normal vector to a surface. In fact, the SDF allows to describe the volumetric surface as the zeroth level-set of $d$, $d(\vect{x})=0$ . The essence of our differential approach is the observation that the surface normal $n$ equals to the gradient of the SDF $d$ as follows 
\begin{equation}
\label{eq:normal_para}
\vect{n}(\vect{x})=\nabla d(\vect{x}).
\end{equation}



Similarly to \cite{Zollhofer15} that used the SDF for single image shading refinement, we consider the SDF for the irradiance equation to derive a differential multi-view photometric stereo formulation where we assume to have $N_{ps}$ images (i.e. light sources) for each known camera position $C_q$ (that is $N_{ps}(C_q)$, $q=1,\ldots N_{views}$).

To exploit the monocular aspect of the photometric stereo problem, we consider image ratios for the Lambertian shading model \cite{lambert1760} assuming calibrated nearby LED light sources
\begin{equation}
\label{eq:irradi}
i_k(\vect{u}(\vect{x}) ) =\rho(\vect{x})a_k(\vect{x})\overline{\vect{n}} (\vect{x})\cdot\overline{\vect{l}}_k(\vect{x})
\end{equation}
where $\vect{u}\in\mathbb{R}^2$ is the image-plane projection of the 3D point $\vect{x}$ and $\rho(\vect{x})$ indicates the albedo. Note that as we are following a volumetric approach, the irradiance equation is considered for a set of 3D points $\vect{x}$.  The bar over a vector means that it is normalized (i.e. $\overline{\vect{n}}=\frac{\vect{n}}{|\vect{n}|}$). 
We model point light sources by considering the following $\vect{l}_k(\vect{x})=\vect{x}-\vect{p}_k$ 
from \cite{Mecca2014near}, where $\vect{p}_k$ is the known position of the point light source with respect to the global coordinate system. We model the light attenuation considering the following non-linear radial model of dissipation
\begin{equation}
\label{eq:attenuation}
a_k(\vect{x})=\phi _k \frac{( \vect{l}_k (\vect{x}) \cdot \vect{s}_k)^{\mu _k}}{||\vect{l}_k (\vect{x})||^2}
\end{equation}
where $\phi _k$ is the intrinsic brightness of the light source, $\vect{s}_k$ is the principal direction (i.e. the orientation of the LED point light source) and $\mu _k$ is an angular dissipation factor.

\vspace{0.2cm}
\noindent \textbf{Modeling with image ratios} As in \cite{Mecca2014near}, we follow the ratio method that significantly simplifies the PS problem by eliminating the dependence on the albedo as well as the non-linear normalisation of the normal. 

Indeed, dividing equations for images $i_h$ and $i_k$ (from the same point of view $C_q$) as in (\ref{eq:irradi}), we have
\begin{equation}
\frac{i_h(\vect{x})}{i_k(\vect{x})} =   \frac{ a_h(\vect{x}) \vect{n}(\vect{x})\cdot\overline{\vect{l}}_h(\vect{x})}{a_k(\vect{x})\vect{n}(\vect{x})\cdot\overline{\vect{l}}_k(\vect{x})}
\end{equation}
which leads to
\begin{equation}
\vect{n}(\vect{x}) \cdot (i_h(\vect{x})a_k(\vect{x})\overline{\vect{l}}_k(\vect{x}) - i_k(\vect{x})a_h(\vect{x})\overline{\vect{l}}_h(\vect{x}) )= 0.
\end{equation}

By substituting the parametrisation of the normal from (\ref{eq:normal_para}), we get the following albedo independent, homogeneous linear PDE
\begin{equation}
\label{eq:fundamental}
\vect{b}_{hk}(\vect{x}) \cdot \nabla d(\vect{x}) = 0 
\end{equation}
where
\begin{equation}
\vect{b}_{hk}(\vect{x})=i_h(\vect{x})a_k(\vect{x})\overline{\vect{l}}_k(\vect{x}) - i_k(\vect{x})a_h(\vect{x})\overline{\vect{l}}_h(\vect{x}).
\end{equation}

The geometrical meaning of (\ref{eq:fundamental}) is the extension to the 3D volumetric reconstruction of the PDE approach presented in \cite{Mecca2014near} and the proposed PS model still consists of a homogeneous linear PDE. 
However, an important difference with \cite{Mecca2014near} is that $\vect{b}_{hk}(\vect{x})$ does not depend on $d$ (i.e. (\ref{eq:fundamental}) is linear and not quasi-linear as proposed in \cite{Mecca2014near}) due to the fact that the relevant quantities are expressed on a global coordinate system independent of the existence of a surface. Crucially, this allows to use the nonlinear lighting model of (\ref{eq:attenuation}) without linearising approximations (e.g. spherical harmonics \cite{Basri2001b}) or dependence on any initial surface estimates.

An interesting observation is that (\ref{eq:fundamental}) is conceptually similar with the iso-depth curves in the work of \cite{Zhou2013multi}. Nonetheless, the SDF formulation is a more natural `object centered' depth and this allows for a unified optimisation as we describe in the next section.

In order to simplify the notation, we will rename the pair $hk$ as $p$ and we will call the set of all the combinations of pairs of images (with no repetition). 

\vspace{0.2cm}
\noindent {\textbf{MVPS as a weighted least squares problem}} With the aim to consider photometric stereo images coming from different views into a single mathematical framework, we stack in a single system the following weighted version of (\ref{eq:fundamental})
\begin{equation}
\label{eq:fundamentalWeighted:1}
w_p(C_q,\vect{x})\vect{b}_p(\vect{x}) \cdot \nabla d(\vect{x}) = 0
\end{equation}
where $w_p(C_q,\vect{x})=\max (\vect{n}(\vect{x})\cdot \vect{v}_q(\vect{x}),0)$. 
This weight term $w_p$ is essentially a measure of visibility. The resulting system then counts $\sum_{q=1}^{N_{views}}\binom{N_{ps}(C_q)}{2}$ equations as shown in (\ref{eq:fundamentalWeighted:2})

\begin{equation}
\label{eq:fundamentalWeighted:2}
\begin{bmatrix}
    [w_1(C_1,\vect{x})\vect{b}_1(\vect{x})]^t \\
    [w_2(C_2,\vect{x})\vect{b}_2(\vect{x})]^t \\
    \vdots        
\end{bmatrix} \nabla d(\vect{x}) = \vect{0}.
\end{equation}

With the aim to solve it as a least square problem, we consider the normal equations
\begin{equation}
\label{eq:lst_fundamental}
B(\vect{x}) \nabla d(\vect{x}) = 0
\end{equation}
with
\begin{equation*}
\begin{split}
B(\vect{x})= [w_1(C_1,\vect{x})\vect{b}_1(\vect{x}),w_2(C_2,\vect{x})\vect{b}_2(\vect{x}), \hdots  ] & \cdot\\ \begin{bmatrix}
    [w_1(C_1,\vect{x})\vect{b}_1(\vect{x})]^t \\
    [w_2(C_2,\vect{x})\vect{b}_2(\vect{x})]^t \\
    \vdots        
\end{bmatrix} &
\end{split}
\end{equation*}
$B(\vect{x})$ is now a positive, semi-definite, 3x3 matrix.

\paragraph{\textbf{B adjustment}} The geometrical constraint coming from (\ref{eq:fundamental}) ensures that all the vector fields $\vect{b}_p(\vect{x})\in\mathbb{R}^3$ span the same bi-dimensional space $\forall \vect{x}$ of the volume as they define the level-set of the SDF. This means that under ideal circumstances, the rank of $B$ in (\ref{eq:lst_fundamental}) should be exactly 2. However, due to numerical approximations this is never exactly true; we enforce this constraint by using eigenvalue decomposition of $B$ hence
\begin{equation}
\label{eq:eigen_b}
B=Q \Lambda Q^{\text{t}}=
 Q \begin{bmatrix}
   \Lambda _1 & 0 & 0\\
   0 &  \Lambda _2 & 0\\    
   0 & 0 & \Lambda _3\\
\end{bmatrix} Q^{\text{t}}
\end{equation}
with
$\Lambda _1 \geq \Lambda _2 \geq \Lambda _3$ and setting $\Lambda _3=0$.

We note that this rank correction is a sanity check step. Indeed if $B \nabla d =0$ with $B$ full rank, then $\nabla d=0$ which can never be true as $|\nabla d|=1$ (Eikonal equation) and so $d$ cannot be the SDF of any real surface. In practice, this would lead to over-smoothing of the SDF  and loss of details. In addition, the Eikonal equation can be implicitly enforced by demanding that 
$\nabla d=\vect{q}_3$, where $\vect{q}_3$ is the third collumn of $Q$ and corresponds to the nullspace of $B$ ($\vect{q}_3$ is a unit vector hence enforcing the Eikonal equation). Hence (\ref{eq:lst_fundamental}) is updated to the following full rank system ($I_{3\times 3}$ is the identity matrix)
\vspace{-0.2cm}
\begin{equation}
\label{eq:lst_fundamental_reg}
\Big(B(\vect{x})+I_{3\times 3}\Big) \nabla d(\vect{x}) =  B'(\vect{x})\nabla d(\vect{x})= \vect{q}_3.
\end{equation}



   \begin{figure}[t]
	\centering   
     \includegraphics[height=0.16\textwidth]{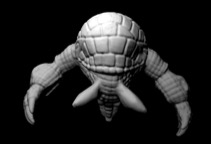}
     \includegraphics[height=0.16\textwidth]{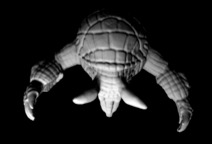}
     \includegraphics[height=0.12\textwidth]{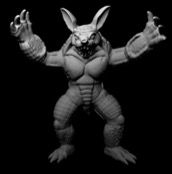}
     \includegraphics[height=0.12\textwidth]{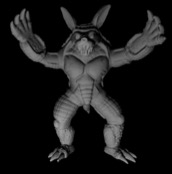}
     \includegraphics[height=0.12\textwidth]{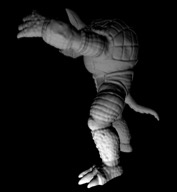}
     \includegraphics[height=0.12\textwidth]{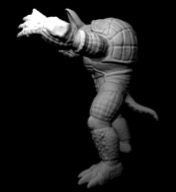}
 \caption{Synthetic data samples where we show per pairs differences in near lighting, perspective deformation and self occlusions.}
    \label{fig:synthetic_data}
  \end{figure}
  
\begin{figure}[b]
\centering   
\subfigure[Ground truth]{\label{fig:synthetic_res_gt}
\includegraphics[width=0.17\textwidth,clip,trim={1.5cm 2cm 1.0cm  1.0cm}]{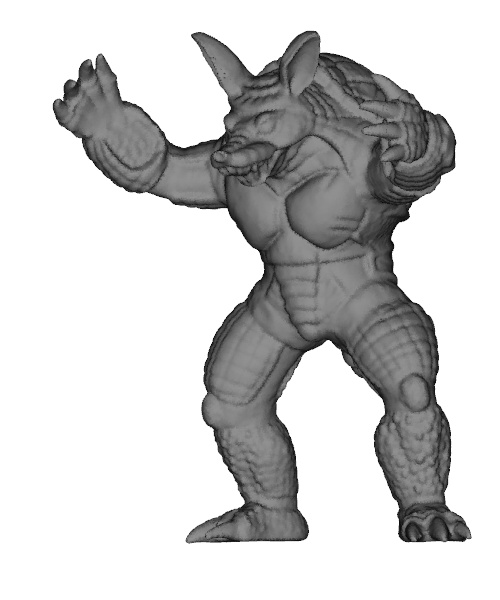}}
\subfigure[Visual Hull]{
\includegraphics[width=0.17\textwidth,clip,trim={1.5cm 2cm 1.0cm  1.0cm}]{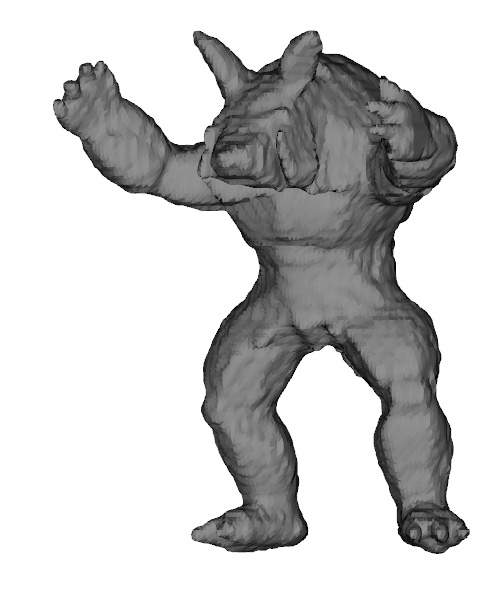}} 
\subfigure[Initial estimates: 500,1500,10K triangles]{
\includegraphics[width=0.15\textwidth,clip,trim={1.5cm 2cm 1.0cm  1.0cm}]{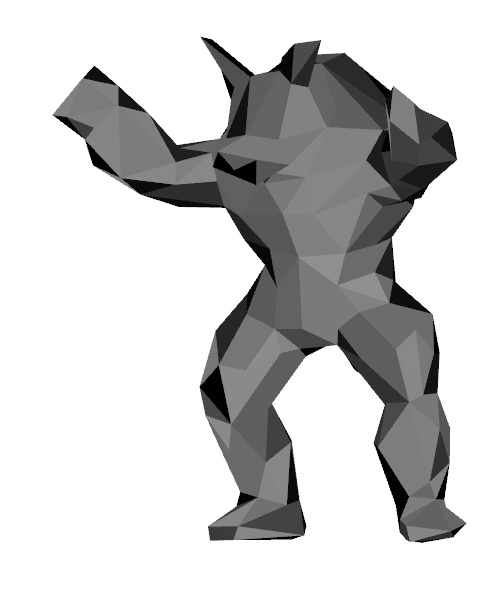}
\includegraphics[width=0.15\textwidth,clip,trim={1.5cm 2cm 1.0cm  1.0cm}]{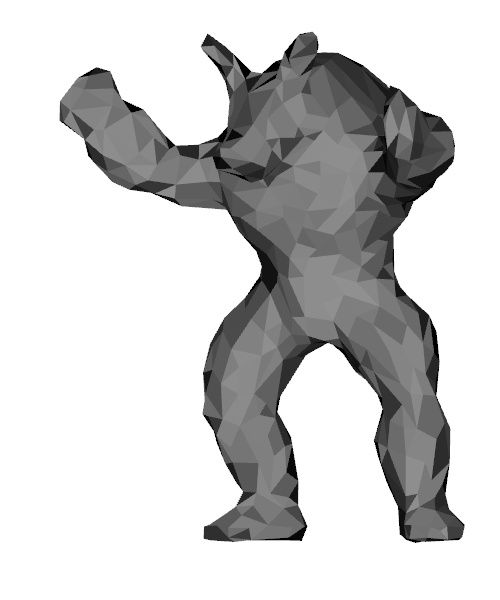}
\includegraphics[width=0.15\textwidth,clip,trim={1.5cm 2cm 1.0cm  1.0cm}]{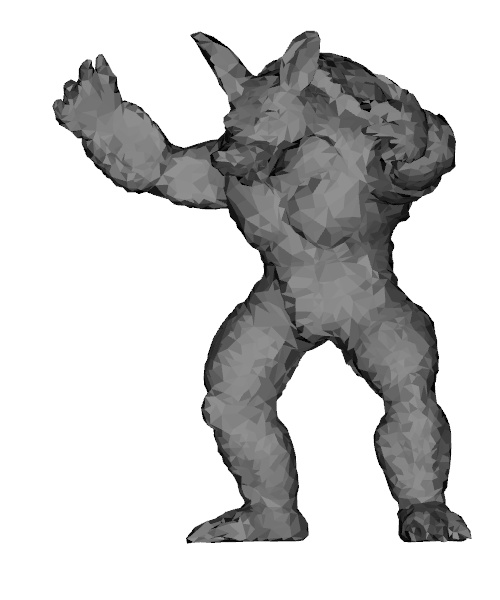}} 
\caption{Synthetic data experiment - initial estimates used for initialising the MVPS optimisation. The variable quality initial estimates of the bottom row are generated by subsampling the ground truth (\ref{fig:synthetic_res_gt}) using Meshlab's  edge collapse decimation function.}
\label{fig:synthetic_meshes}
\end{figure}
  
  \begin{figure*}[h]
	\centering
	\includegraphics[width=0.0275\textwidth]{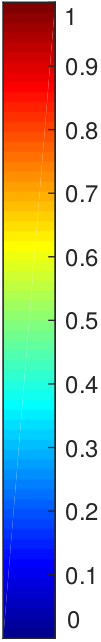}
\subfigure[\cite{Park2017}-RMS Err 0.105mm]{\label{fig:arma_pami_gt}
\includegraphics[width=0.228\textwidth,clip,trim={4.5cm 9.5cm 2.0cm  2.0cm}]{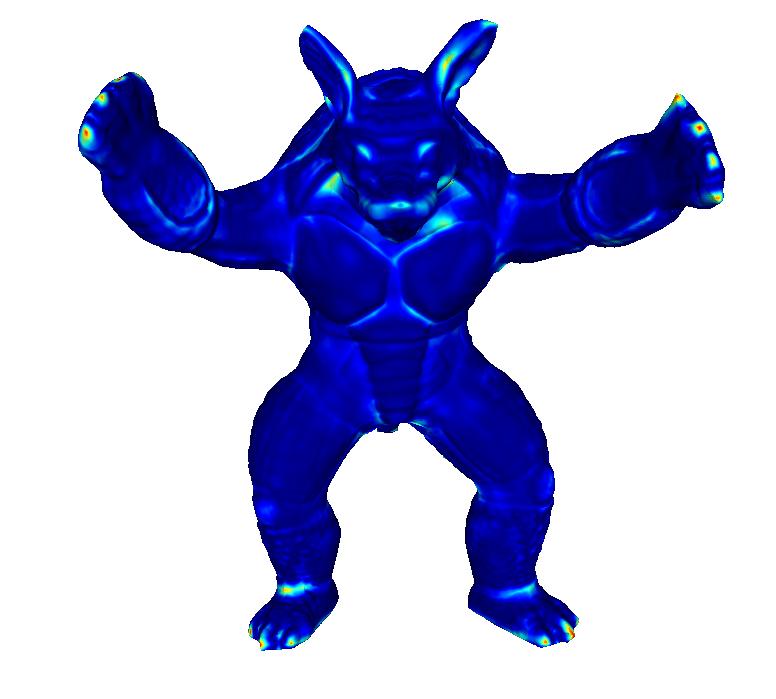}} 
\subfigure[Ours-RMS Err 0.090mm]{\label{fig:arma_from_gt}
\includegraphics[width=0.228\textwidth,clip,trim={4.5cm 9.5cm 2.0cm  2.0cm}]{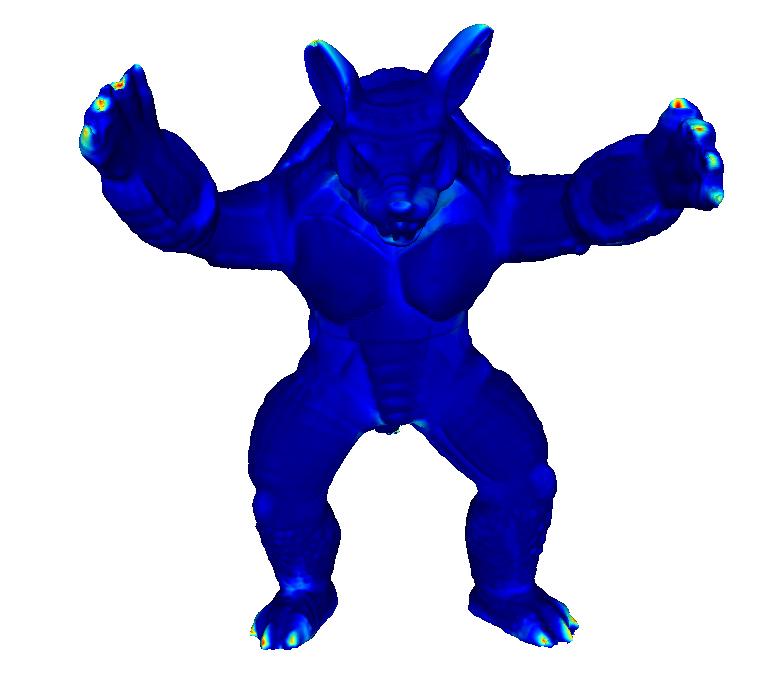}} 
\subfigure[\cite{Park2017}-RMS Err 0.370mm]{\label{fig:arma_pami_vh}
\includegraphics[width=0.228\textwidth,clip,trim={ 6.75cm 5.5cm 5.25cm  1.25cm}]{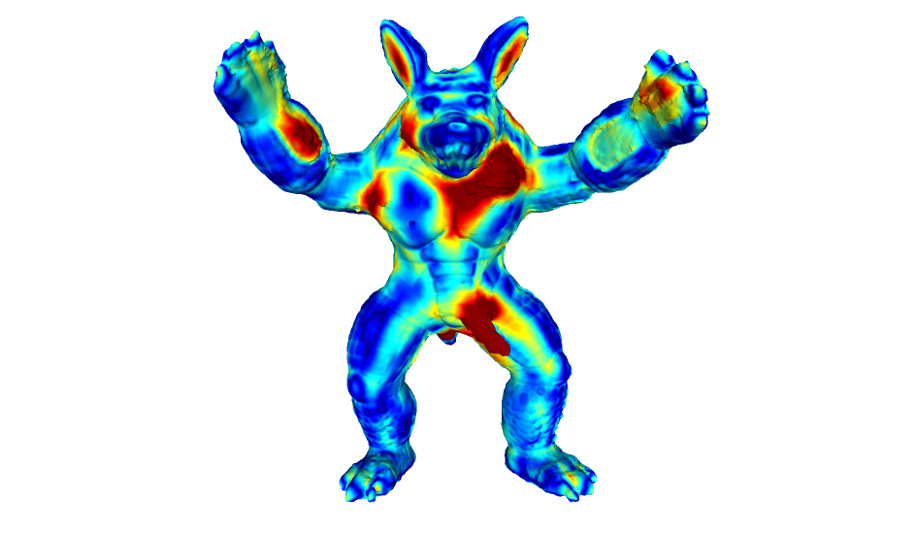}} 
\subfigure[Ours-RMS Err 0.293mm]{\label{fig:arma_from_vh}
\includegraphics[width=0.228\textwidth, clip,trim={ 6.75cm 5.5cm 5.25cm  1.25cm}]{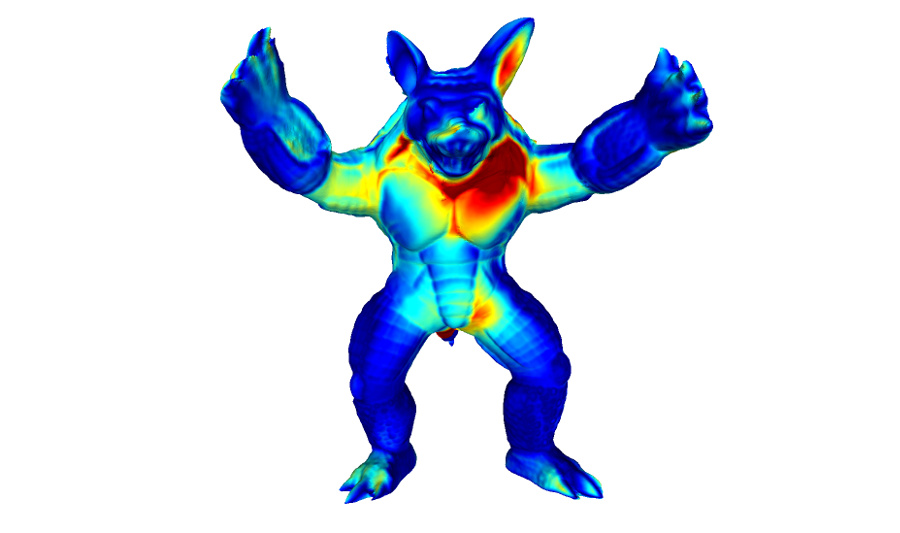}}
 \caption{Evaluation using the 1500 triangles mesh initial estimate \subref{fig:arma_pami_gt},\subref{fig:arma_from_gt}  and the visual hull initial estimate \subref{fig:arma_pami_vh},\subref{fig:arma_from_vh}. The colour coding shows the error in millimeters compared to the ground truth (computed using Meshlab's Hausdorff distance function).}
    \label{fig:synthetic_data_res}
  \end{figure*}

    \section{Variational resolution}
\label{sec:numericalPart}

In this section, we describe how we build the variational solver to compute the signed distance field based parameterisation introduced in the previous section.
\vspace{-0.35cm}
\paragraph{\textbf{Discretisation}} To avoid excessive computation, we note that the photometric stereo equations do not need to be computed in the whole volume but rather only to a subset of voxels $\Omega \subset \mathbb{R}^3$, which are close to the surface. In fact, (\ref{eq:normal_para}) is only true in the vicinity of the surface. We discretise the variational problem (\ref{eq:lst_fundamental_reg}) by using first order forward finite differences $\nabla \vect{d}= G \cdot \vect{d}$, with $\vect{d}$ being the vector stacking $d(\vect{x})$ in all of $\Omega$. $G$ is the sparse kernel matrix describing the connectivity in $\Omega$. 

We note that the differential approach is inevitably rank deficient (needing an integration constant for the theoretical minimal solution).
Hence, we follow the standard of most modern variational approaches (e.g. \cite{MeccaQLC2016}) and adopt a Tikhonov regulariser of the form $\vect{d}=\vect{d}_0$ where $\vect{d}_0$ is some initial estimate of the SDF obtained from the distance transform of the an initial surface estimate. Also note that differential approaches are guaranteed to recover  smooth surfaces without the need of a smoothness regularisers (like the one of \cite{GuoTOG2017}).
Thus the problem becomes (using $\lambda =0.05$)
\begin{equation}
\label{eq:optim_full}
\min_{\vect{d}} ( || B_{\Omega} G \vect{d}- \vect{q}_{\Omega} || +\lambda ||\vect{d}-\vect{d}_0||)
\end{equation}
where $B_{\Omega}$ and $ \vect{q}_{\Omega}$ are obtained by stacking the relevant quantities from (\ref{eq:lst_fundamental_reg}) for all voxels in $\Omega$. The resulting linear system is solved with the conjugate gradients method. Jacobi preconditioner is used as the system is too large ($10^7-10^9$ elements) to use a more sophisticated one.

\subsection{Octree Implementation}

To manage the required set of voxels $ \in \Omega$ described above we use an octree structure. 
$\Omega$ is defined at the leafs of the tree and Voxel neighbors for computing finite differences are found by bottom up traversal of the tree.

We perform an iterative procedure of solving (\ref{eq:optim_full}) on the leafs of the tree and then subsequently subdividing those leafs where the absolute value of SDF is smaller than 2 voxel sizes.  
The procedure repeats until the voxels are small enough so as their projection on the image planes is smaller than the pixel size and thus the maximum obtainable resolution has been reached. As a result, only a small fraction of the volume is considered for calculations and the hierarchy of voxels is densely packed around the surface. Finally, the reconstructed surface is computed with the Marching cubes variant of \cite{kazhdan2007unconstrained}. 

It is important to note however that this iterative procedure is only needed for computational reasons. If the whole volume could be filled with voxels, solving (\ref{eq:optim_full}) would recover the whole surface in a single step.

\begin{table}[b]
\caption {Quantitative evaluation based on the initial estimate quality. Errors are in mm. Noise added to vertex positions and the magnitude is relative to the average triangle size.} 
\label{tab:quality_comp} 

\begin{center}
\scalebox{0.75}{
\begin{tabular}{ |c|c|c|c|c|c|c|c|c| } 
\hline
\multicolumn{2}{|c|}{Experiment} &  \multicolumn{5}{|c|}{Triangle Number} & Visual Hull\\
\hline
Method & Noise & 250 & 500 & 1500& 10K & 30K & 14K\\
\hline
\multirow{3}{4em}{\cite{Park2017}} & 0\% & 0.245 & 0.141 & 0.105 & 0.029 & 0.025 & 0.370\\ 
& 5\% & 0.290 & 0.172 & 0.119 & 0.036 & 0.029 & -\\ 
& 10\%& 0.393 & 0.250 & 0.153 & 0.046 & 0.031 & -\\ 
\hline
\multirow{3}{4em}{Proposed} & 0\% & 0.203 & 0.114 & 0.090 & 0.026 & 0.023 & 0.293\\ 
& 5\% & 0.234 & 0.137 & 0.104 & 0.033 & 0.024 & -\\ 
& 10\%& 0.321 & 0.193 & 0.131 & 0.043 & 0.028 & -\\ 
\hline
\end{tabular}
}
\end{center}
\end{table}
\vspace{-0.4cm}
\paragraph{\textbf{Visibility Estimation}} In order to deal with scenes with a complex geometry and multiple objects, occlusions and cast shadows need to be addressed. This is performed by ray-tracing lines from each voxel to each light source and camera and using the current estimate of the geometry. 
As it is well accepted in the graphics community (e.g \cite{luebke2003level}), octree structures allows for very quick visibility checks and 
whenever an occlusion/shadow is detected, the relevant weight in (\ref{eq:fundamentalWeighted:2}) is set to 0. 

The details of the tree evolution and raytracing operations are elaborated in the supplementary \S A.

The method was implemented in Matlab with the octree code using c++ in a mex file.
\vspace{-0.3cm}
\section{Experiments}
\label{sec:numericalTest}
With the aim to prove the capability of our approach to reconstruct complex scenes, we considered both synthetic and real data. We compared against \cite{Park2017} using the code from their website. It is worth to mention that differently from our method, their state-of-the-art approach for MVPS is based on a fully un-calibrated PS model.

For the synthetic case, we used the Armadillo model from the Stanford 3D Scanning Repository\footnote{http://graphics.stanford.edu/data/3Dscanrep/}. The virtual object was scaled to have approximate radius 20mm  and the virtual camera of focal length 6mm was placed in several locations on a sphere of 45mm around the object. We rendered 12 views with 8 images each of resolution 1200x800x24bits per pixel (see Figure~\ref{fig:synthetic_data}).

\begin{figure}[b]
	\centering 
\includegraphics[width=0.3\textwidth]{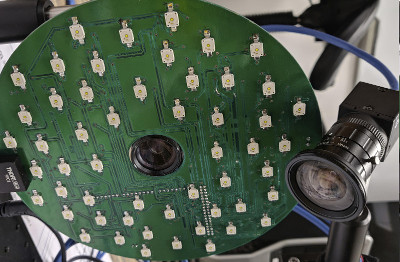}
 \caption{Hardware setup used for the acquisition.  Only 9/52 LEDs are used for the acquisitions. The stereo configuration allows to get a wider view from the camera on the right whereas the camera in the center of the PCB is able to acquire near-field views of the objects.} 
    \label{fig:setup}
  \end{figure}

In order to quantitatively describe the dependency of the accuracy of the volumetric reconstruction to the initial estimate, we subsampled the initial mesh\footnote{Using the quadric edge collapse decimation function of Meshlab.} to 5 different meshes with number of triangle ranging from 250 to 30K (the original mesh was 150k triangles). For each of these meshes we added Gaussian noise to the vertex coordinates with std 0, 5, 10\% of the average triangle size. 
The visual hull (computed with voxel carving) was also used for a final experiment.
 
The evaluation metric is the RMS Hausdorff distance to the ground truth (computed with Meshlab). Results are shown in Figures~\ref{fig:synthetic_data_res} and Table~\ref{tab:quality_comp}. The proposed approach outperforms \cite{Park2017} in all experiments with the difference being more significant on the low quality initial estimate experiments. 


  \subsection{Real Data}
  \label{subsec:realdata}
\vspace{-0.1cm}
For acquiring real world data we used an active light system (see Figure \ref{fig:setup}) consisting of two FLIR cameras FL3-U3-32S2C-CS. One camera mounted an 8mm lens and was surrounded by OSRAM ultra bright LEDs for capturing data in the near-field. The second camera had a 4mm lens for acquiring a wider area helping to track the trajectories of both cameras. 
The stereo pair was calibrated using Matlab's computer vision toolbox and a checkrboard, in order to be able to capture the absolute scale of the world, which is needed for the near light modeling (Equation~(\ref{eq:attenuation})).

 \begin{figure}[t]
 \begin{center}
 \includegraphics[height=0.24\textwidth]{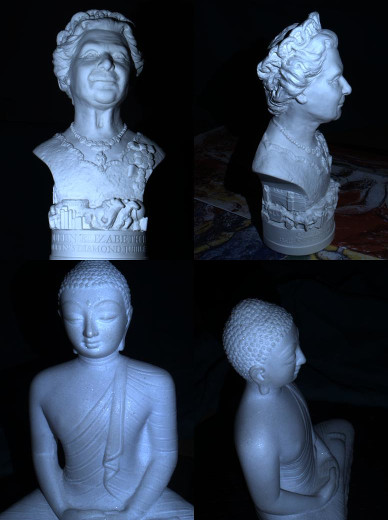} 
 \includegraphics[height=0.24\textwidth, clip,trim={2.25cm 1.5cm 1.5cm  0.0cm}]{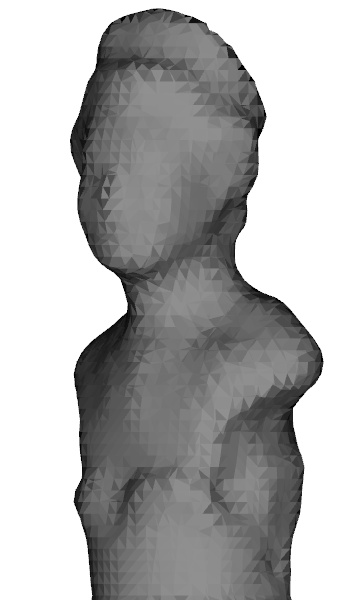}  
 \includegraphics[height=0.24\textwidth, clip,trim={0.25cm 0.5cm 1.75cm  2.0cm}]{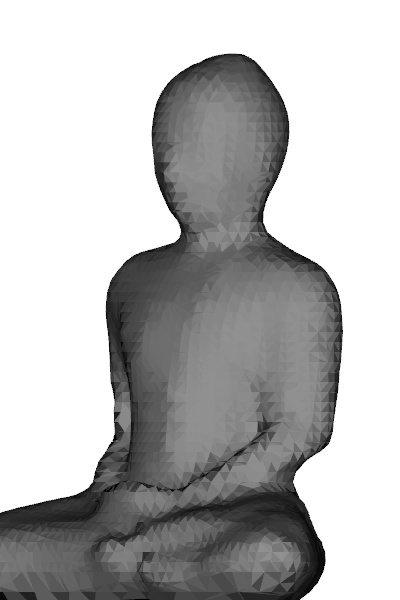}
 \end{center}
 \caption{Real data: 2/108 photometric stereo images (we used 12 views with 9 lights in each view) and initial geometry estimate obtained with MVS. This initial estimates are only 8k and 11k triangles for the Queen and Buddha datasets respectively.}
  \label{fig:real_data_1}
  \end{figure}
  
\begin{figure}[b]
	\centering 
\includegraphics[height=0.14\textheight,clip,trim={1.5cm 0.0cm 1.5cm  0.0cm}]{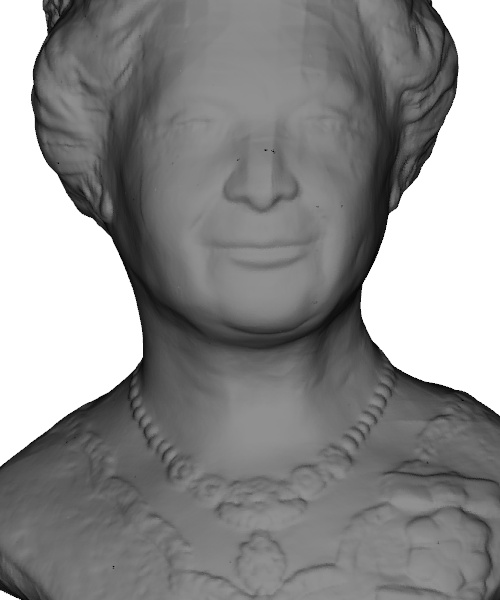} 
\includegraphics[height=0.14\textheight,clip,trim={1.5cm 0.0cm 1.5cm  0.0cm}]{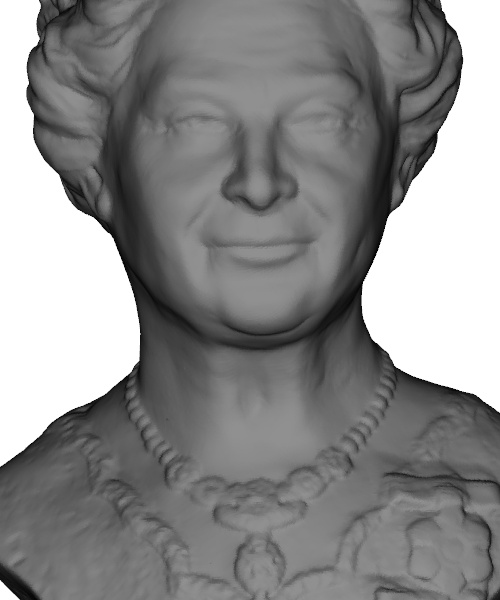}
\includegraphics[height=0.155\textheight,clip,trim={1.5cm 0.0cm 1.0cm  0.0cm}]{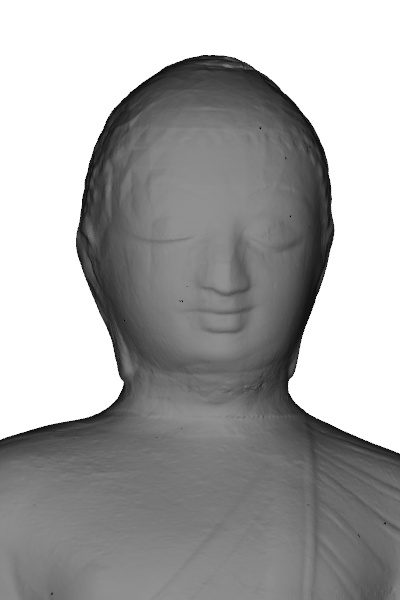}
\includegraphics[height=0.145\textheight,clip,trim={1.5cm 0.0cm 1.0cm  0.0cm}]{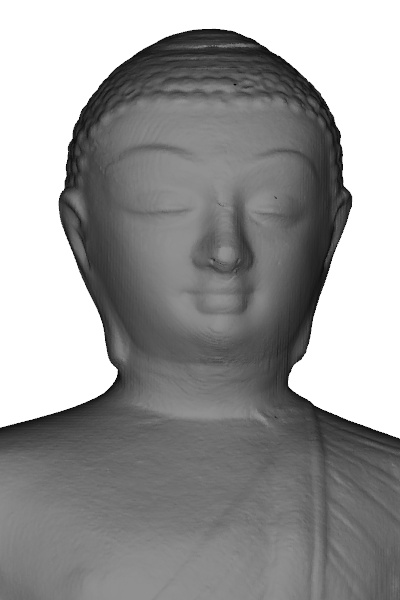}
\caption{Qualitative evaluation on real data set of Figure~\ref{fig:real_data_1}. The proposed approach outperforms \cite{Park2017} and generates more detailed reconstructions.
 }
    \label{fig:real_results}
  \end{figure}

The images have been acquired while moving the setup around the scene. We used COLMAP-SFM  \cite{schoenberger2016sfm,schoenberger2016mvs} to process multi-view data to get camera rotation and translation between the photometric stereo views as well as a low quality reconstruction to use as initial estimate. In addition, a few more images were captured in between the photometric stereo sequences (with neutral illumination) in order to make SFM more robust with respect to a too small overlap between images. To make the models obtained through MVS have less noise, we remove some noisy regions and background points far away from the scenes of interest. Then, we performed Poisson reconstruction \cite{kazhdan2006poisson} with a low level setting so as the initial estimate contains continues surfaces (and not point clouds). As Table~\ref{tab:quality_comp} suggests, our method does not need a very accurate initial estimate. Finally, the initial SFD $\vect{d}_0$ is computed as the distance transform of the initial surface. 

We performed minimal prepossessing to remove saturated and almost saturated pixels as they likely correspond to specular highlights, which are inconsistent with the Lambertian shading model assumed.

\begin{figure}[t]
\centering  
\includegraphics[clip,width=0.235\textwidth,trim={10.5cm 5.5cm 18.5cm  0.0cm}]{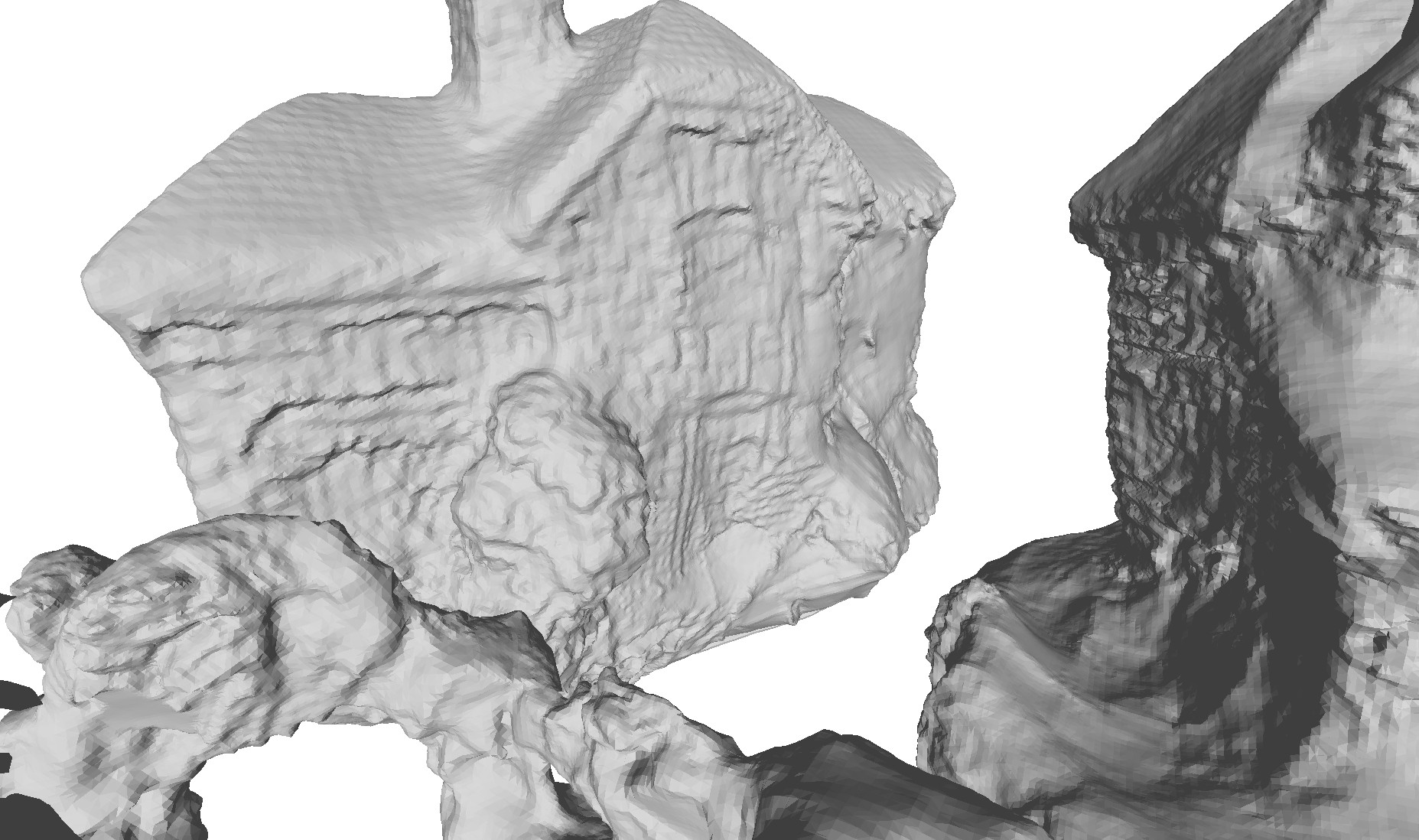}
\includegraphics[clip,width=0.235\textwidth,trim={10.5cm 5.5cm 18.5cm  0.0cm}]{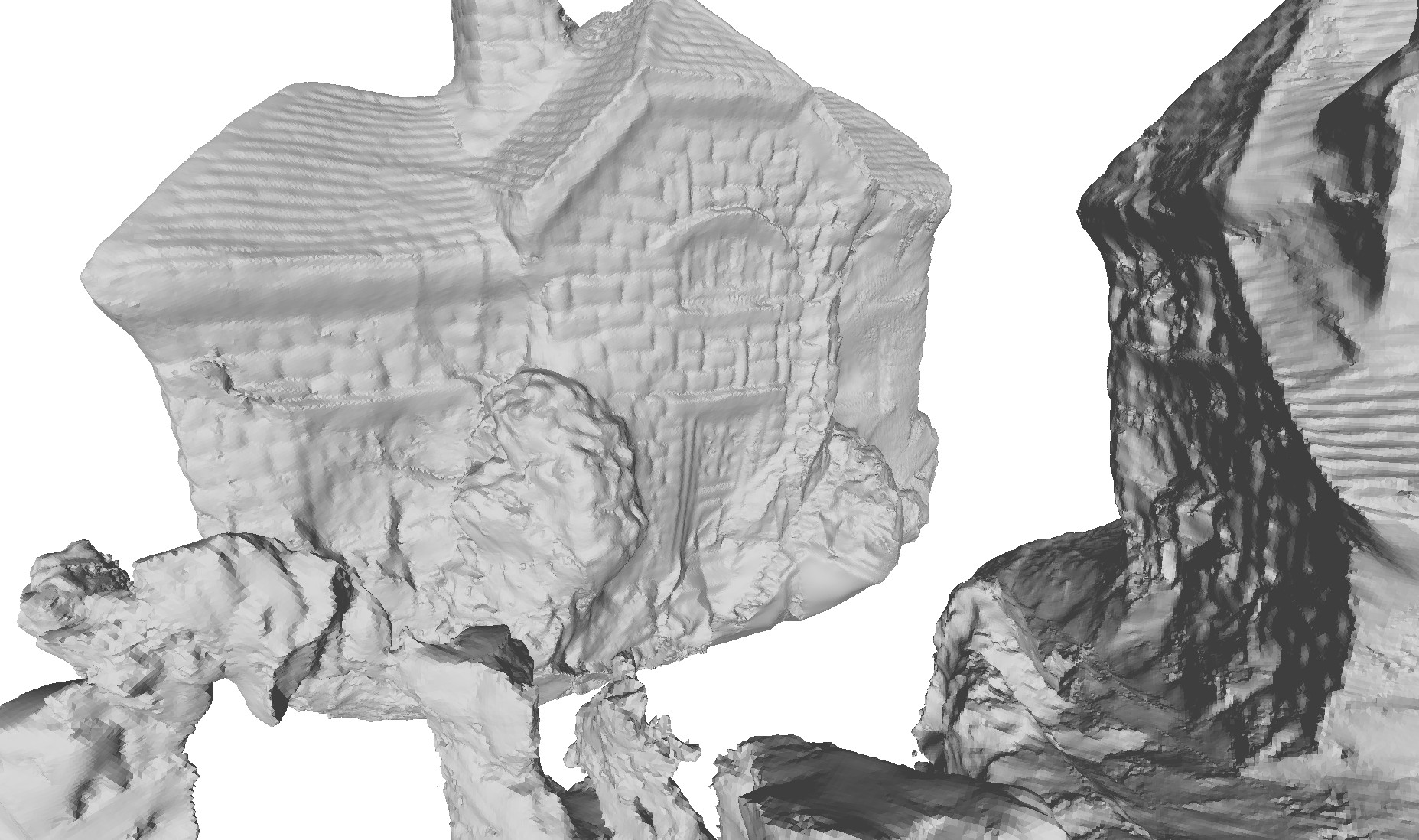}
\includegraphics[clip,width=0.235\textwidth,trim={13cm 1.5cm 21.0cm  0.5cm}]{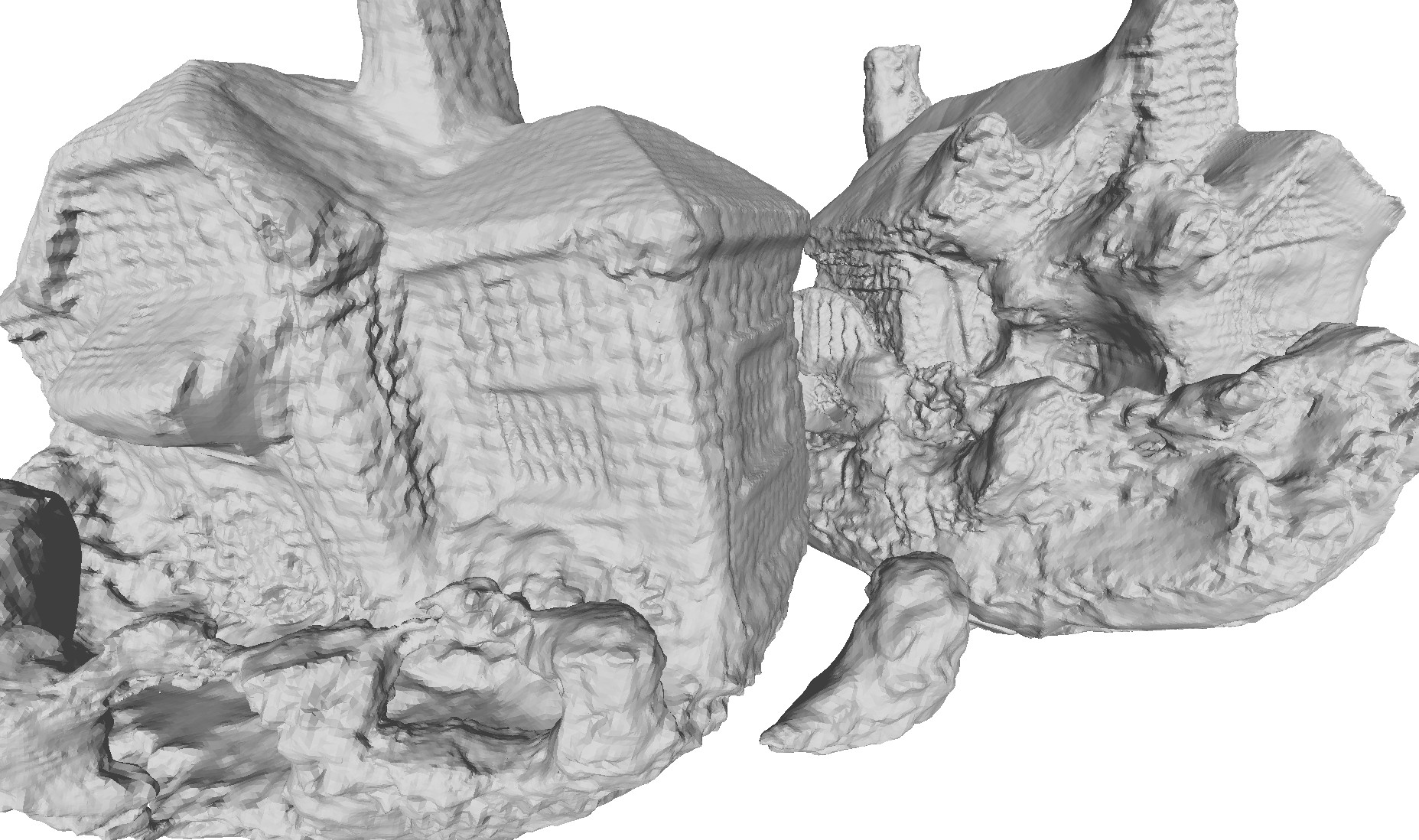}
\includegraphics[clip,width=0.235\textwidth,trim={13cm 1.5cm 21.0cm  0.5cm}]{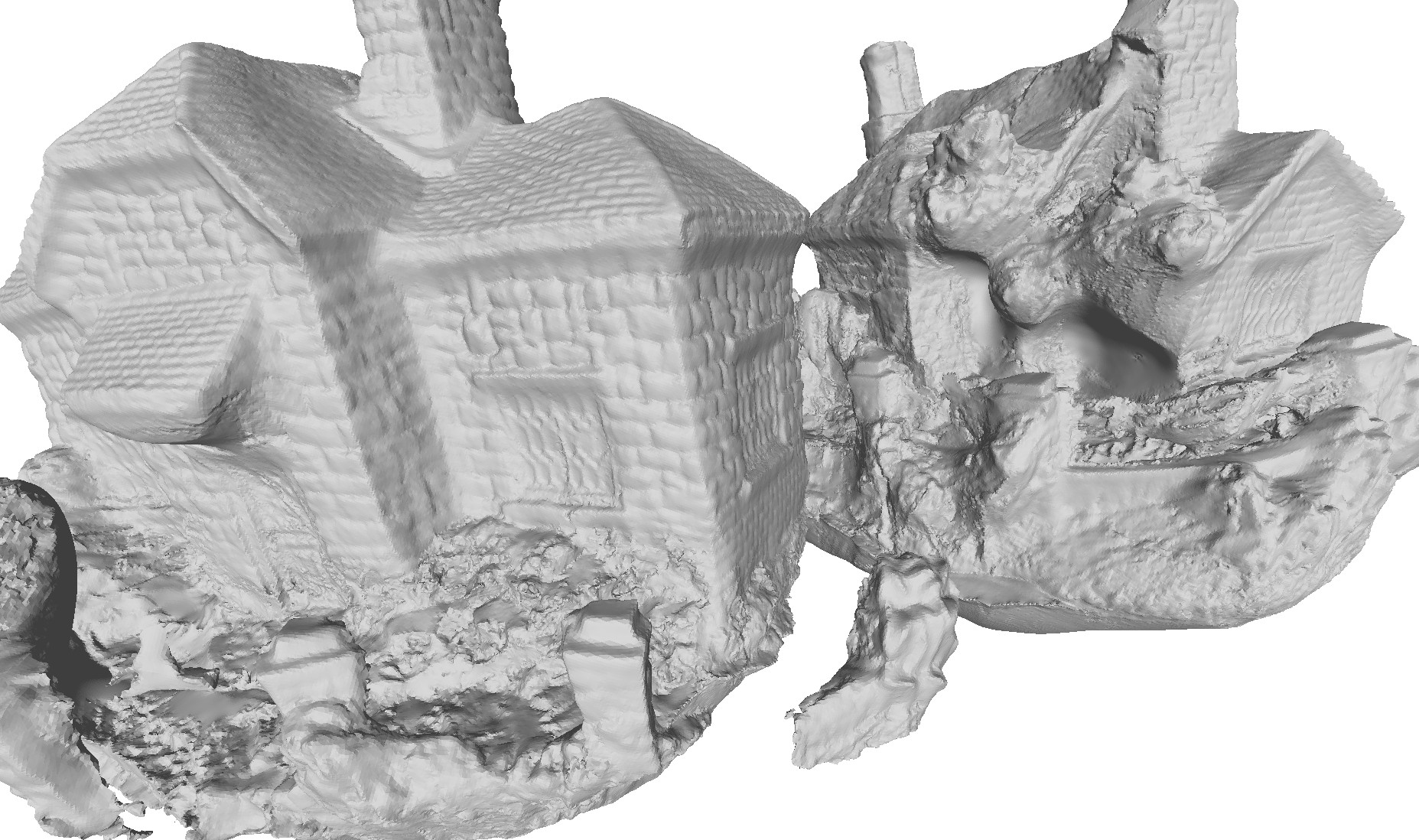}
\caption{Closeup rednerings showing comparison of \cite{Park2017} (\textit{left}) to the proposed  method (\textit{right}). Note that the  proposed method is clearly outperforming the competition, especially in the top row which corresponds to a view in the middle of the scene which is particularly challenging due to cast shadows (see Figure~\ref{fig:real_data_houses_intro}).}
    \label{fig:house_comp}
  \end{figure}

   \begin{figure*}[ht]
	\centering  
    \includegraphics[clip,width=0.33\textwidth,trim={16.0cm 2.5cm 15cm  1.0cm}]{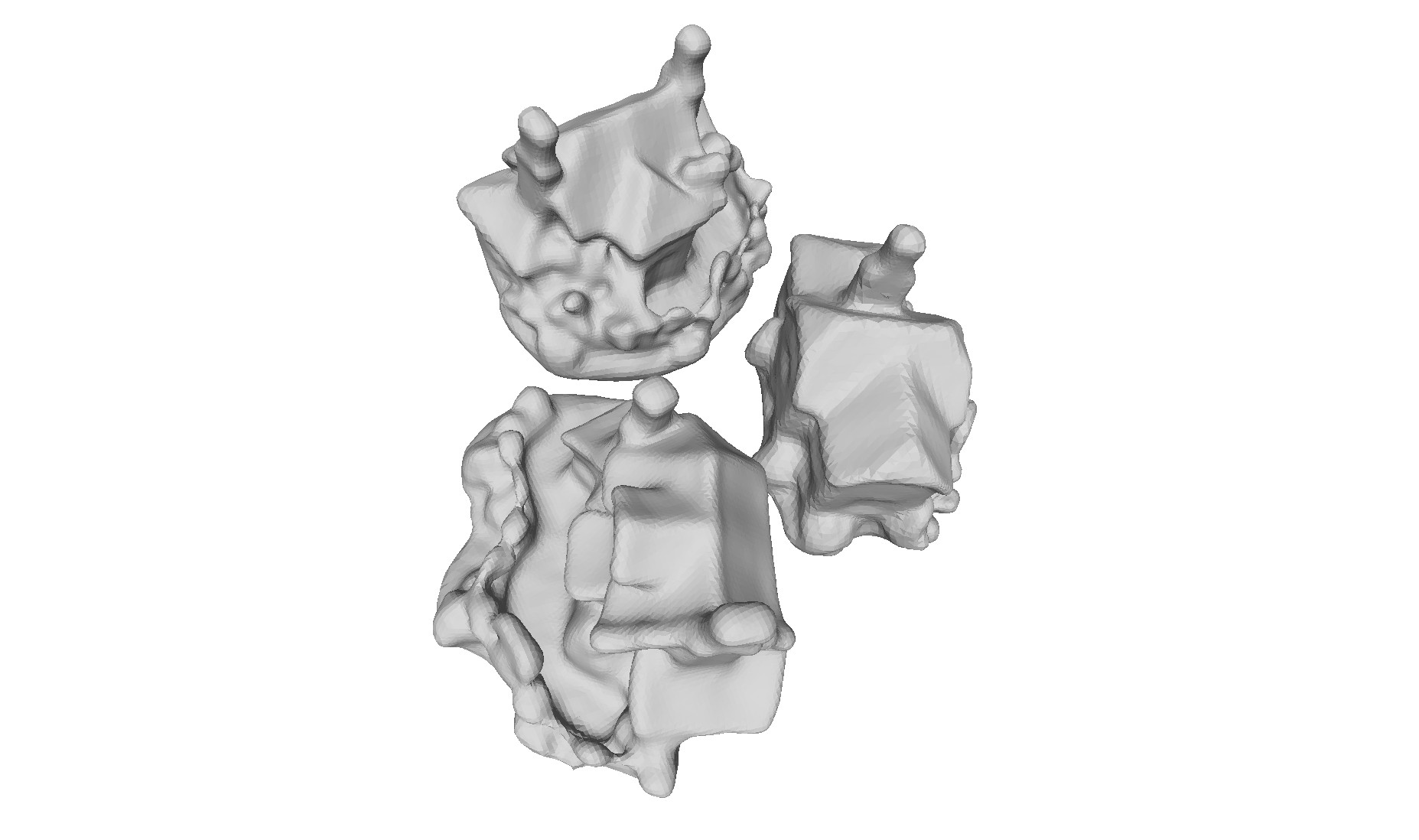}
   \includegraphics[clip,width=0.33\textwidth,trim={16.0cm 2.5cm 15cm  1.0cm}]{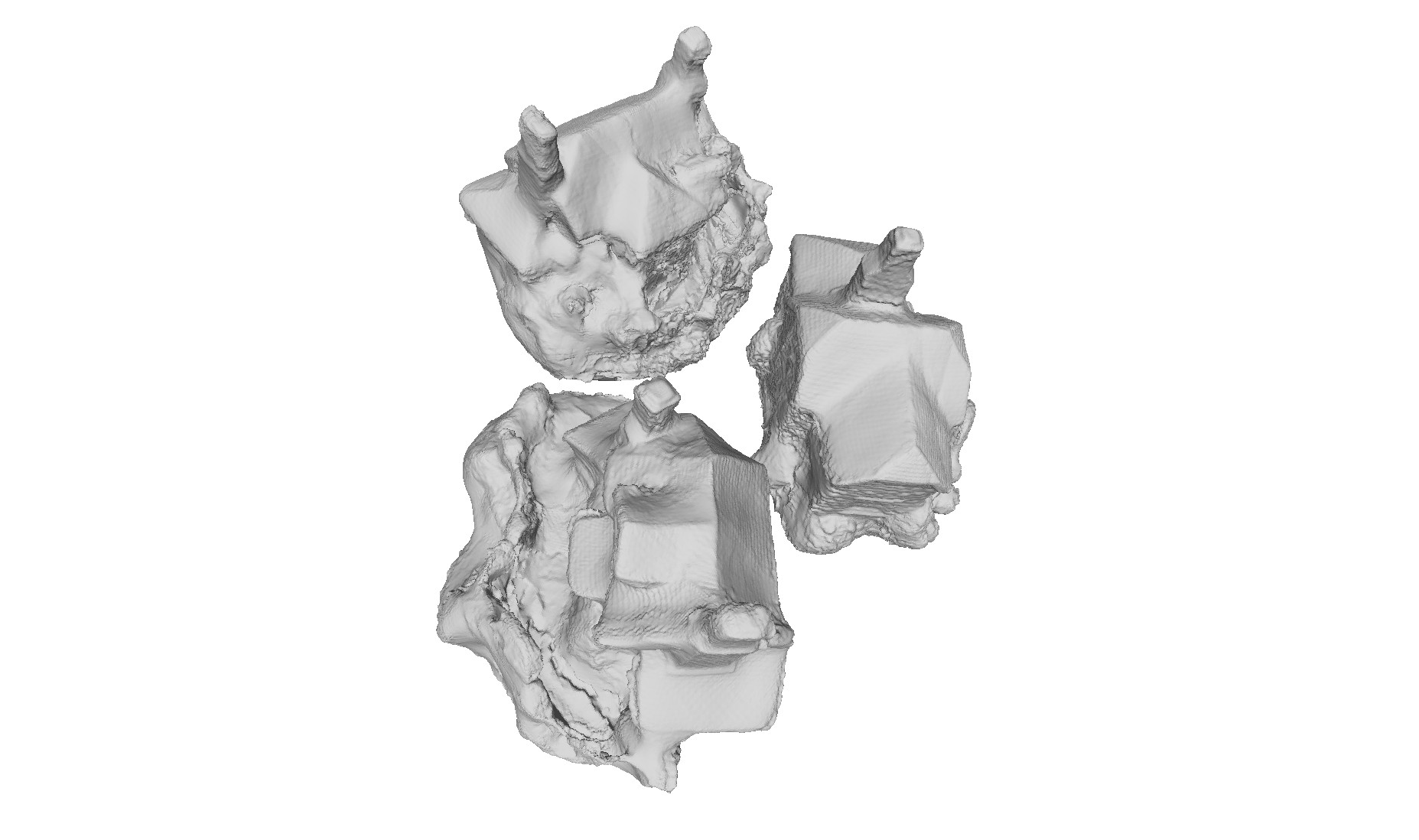}
   \includegraphics[clip,width=0.33\textwidth,trim={16.0cm 2.5cm 15cm  1.0cm}]{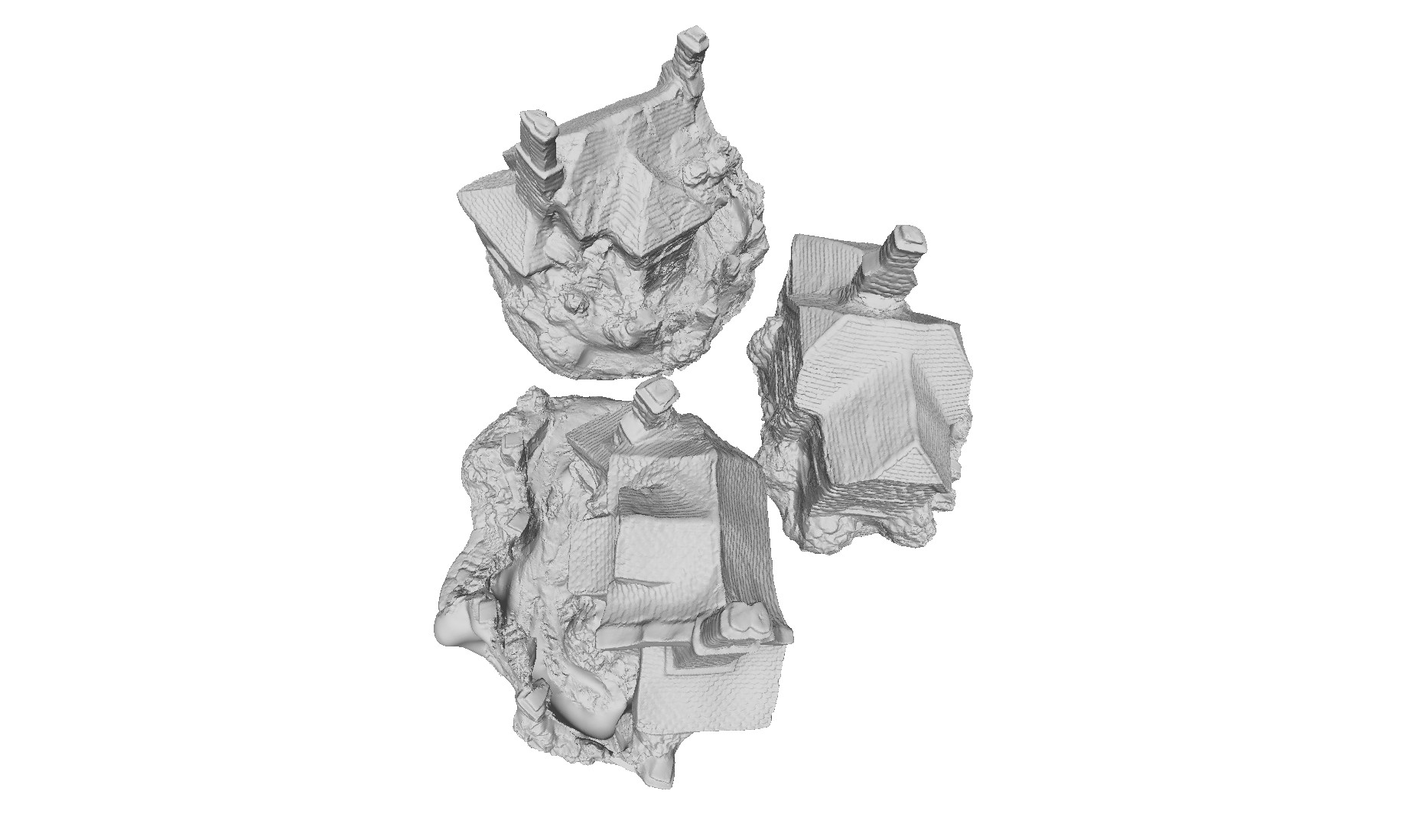}
 \caption{Top view of the reconstruction of the `village' dataset presented in Figure~\ref{fig:real_data_houses_intro}. This includes the initial geometry estimate obtained with SFM(\cite{schoenberger2016sfm,schoenberger2016mvs} (\textit{left}),  \cite{Park2017} (\textit{middle}) and the proposed method  (\textit{right}). }
    \label{fig:house_top}
  \end{figure*}

\begin{figure}[b]
	\centering  
\includegraphics[clip,width=0.152\textwidth,trim={0cm 0cm 0cm  0cm}]{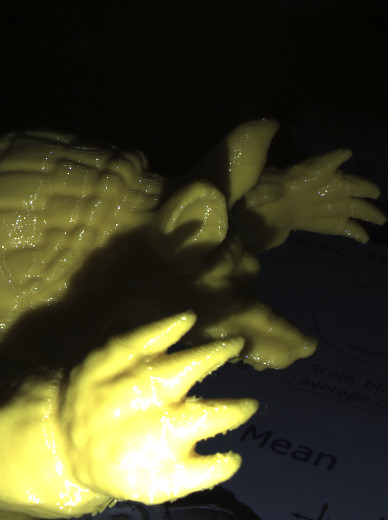}
\includegraphics[clip,width=0.152\textwidth,trim={0cm 0cm 0cm  0cm}]{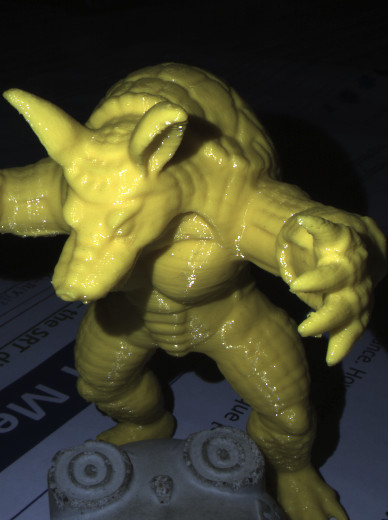}
\includegraphics[clip,width=0.152\textwidth,trim={0cm 0cm 0cm  0cm}]{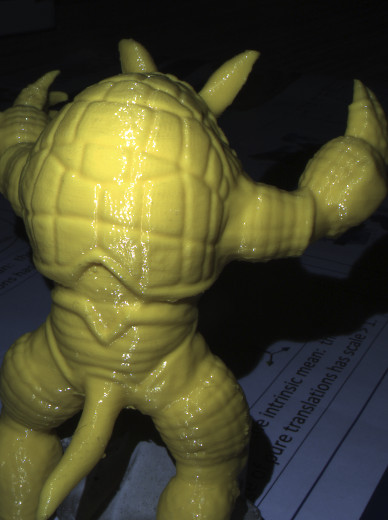}
 \caption{Sample images (3/90) from the Armadillo-owl dataset. The height of the armadillo is around 15cms. }
    \label{fig:owl_arma_samples} 
  \end{figure}

Our real-world datasets include a marble Buddha statue, plaster bust of Queen Elisabeth (Figure~\ref{fig:real_data_1}), a plastic 3D printed version of the Armadillo next to an owl statue (Figure~\ref{fig:owl_arma_samples}) as well as toy village scene (Figure~\ref{fig:real_data_houses_intro}).

 The experiments were run on a server machine. The computational requirements were 15-20GB of RAM and 10-20 minutes of CPU time for the synthetic experiments and the single object datasets showed in Figure \ref{fig:real_data_1}. Figure \ref{fig:real_data_houses_intro} and \ref{fig:owl_arma_samples} correspond to much bigger volumes and thus they needed around 2 hours and 150GB of RAM; this cost is justified by the fact that the octree was allowed to grow to around 30M voxels of size 0.2mm which is required to cover the few litters volume.

The proposed approach outperforms \cite{Park2017} in all datasets. In the simple objects (see Figures~ \ref{fig:real_results}) the difference is that our method is able to get more detailed surfaces as it is not limited by the initial estimate and iteratively refines the surface to the limit of precision. This is in contrast to \cite{Park2017} which is limited by the initial estimates as it is used to create a 2D domain and the calculate a displacement map. Note that using \cite{Park2017} in an iterative fashion is explicitly discouraged by the authors as the 2D parameterisation gets very expensive if the initial estimate has a high triangle count. In addition, their 2D domain is set to a fixed resolution and thus cannot generate surfaces with arbitrary level of details. This is in contrast to our octree implementation which naturally allows continues refinement and different level of precision in different parts of the scene.

       \begin{figure}[b]
	\centering  
    \includegraphics[clip,width=0.235\textwidth,trim={8.0cm 3.5cm 16.0cm  1.5cm}]{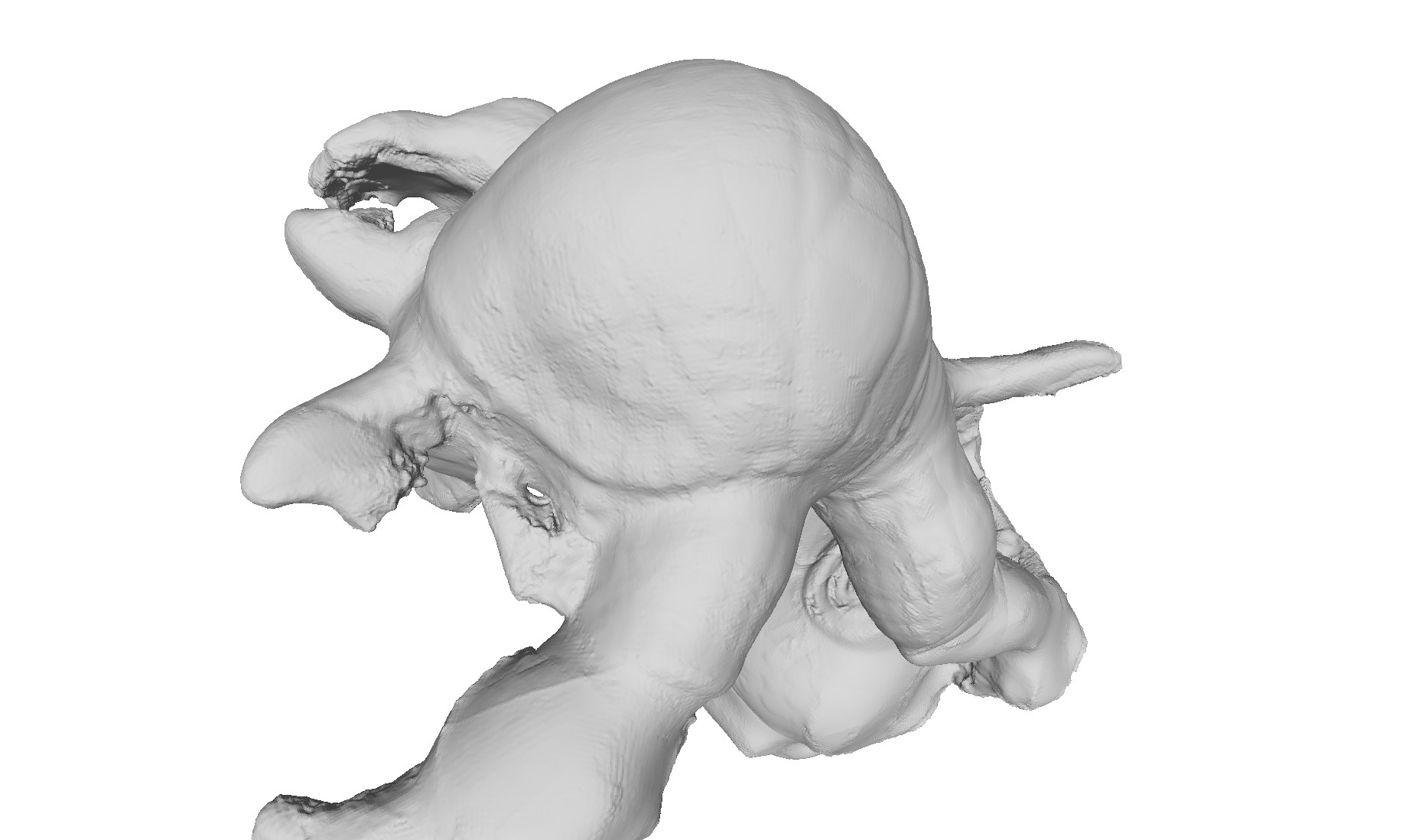}
   \includegraphics[clip,width=0.235\textwidth,trim={8.0cm 3.5cm 16.0cm  1.5cm}]{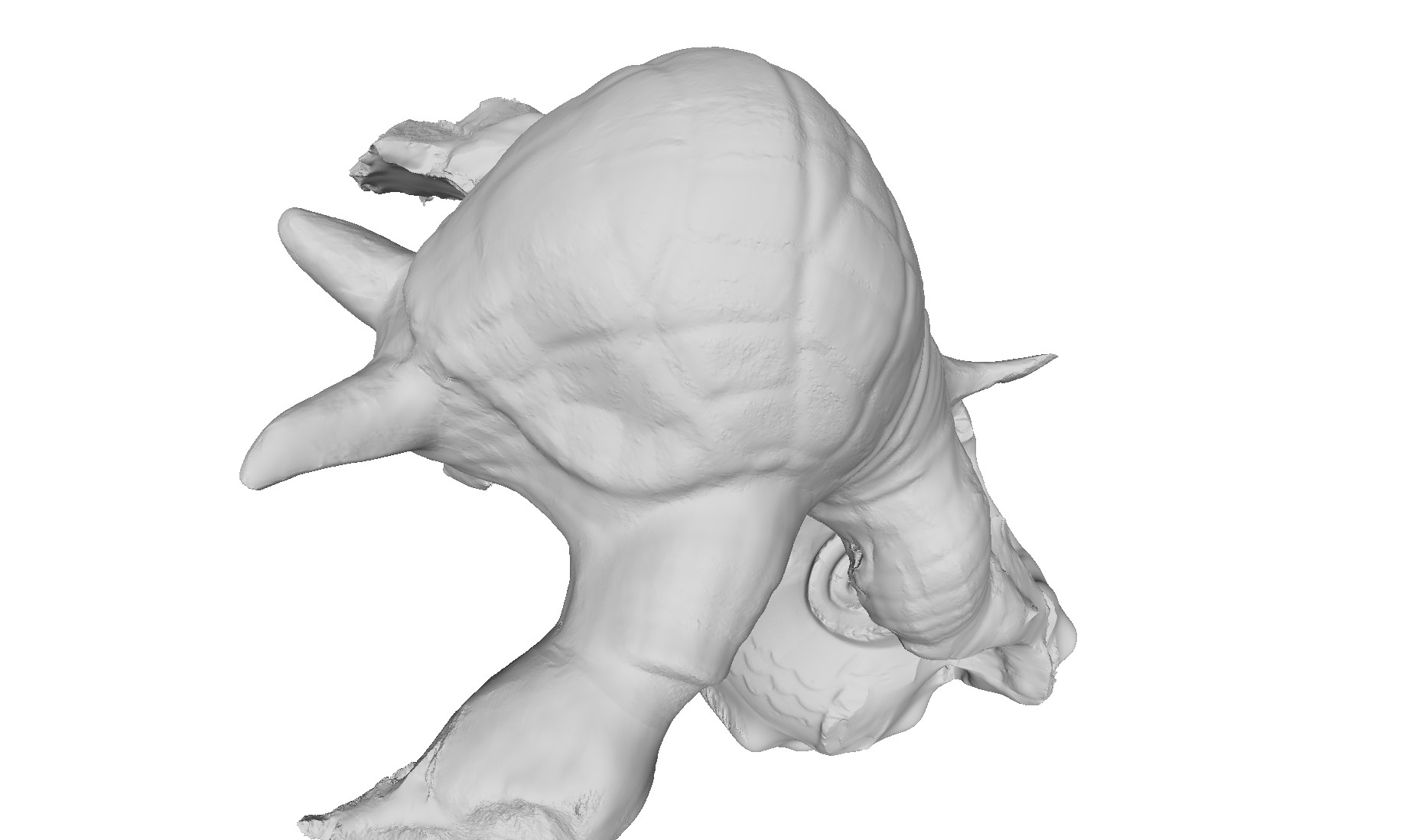}
   \includegraphics[clip,width=0.235\textwidth,trim={12cm 5.5cm 15.0cm  10.0cm}]{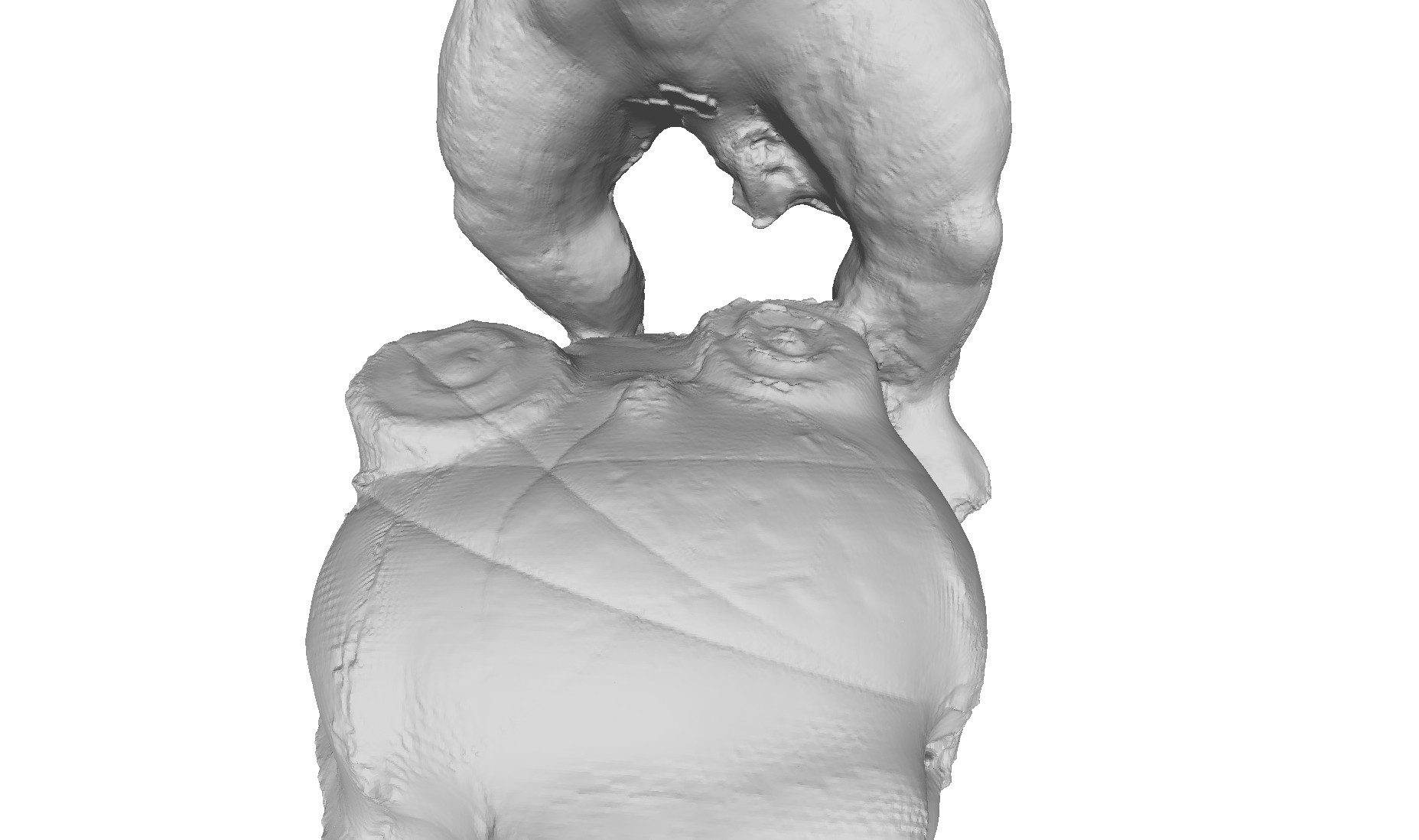}
   \includegraphics[clip,width=0.235\textwidth,trim={12cm 5.5cm 15.0cm  10.0cm}]{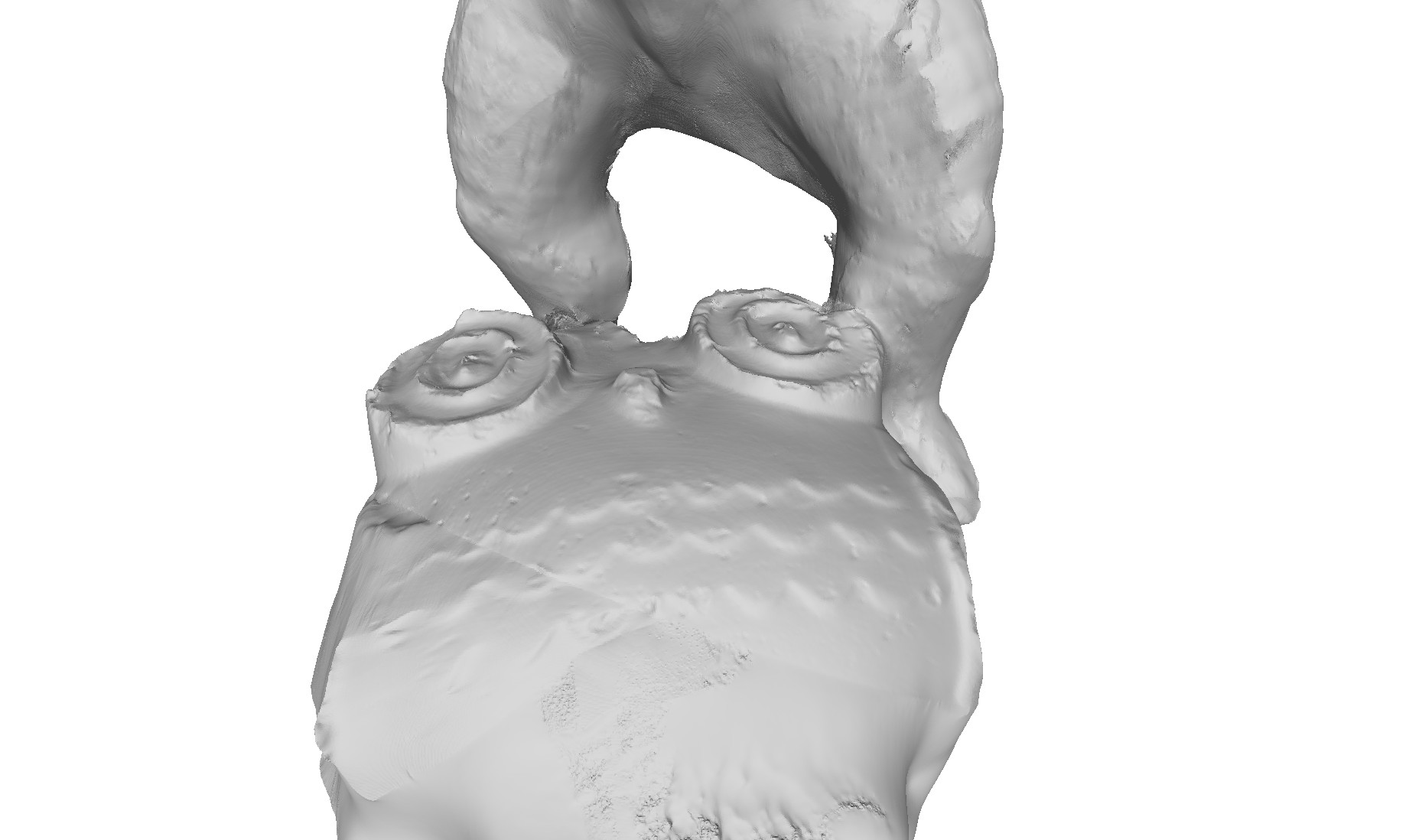}
 \caption{Closeup rednerings showing comparison of \cite{Park2017} (\textit{left}) to the proposed  method  (\textit{right}) for the dataset of Figure~\ref{fig:owl_arma_samples}. }
    \label{fig:owl_arma_comp}
  \end{figure}
  
       \begin{figure*}[t]
	\centering  
    \includegraphics[clip,width=0.32\textwidth,trim={14.0cm 4.5cm 17.5cm  2.0cm}]{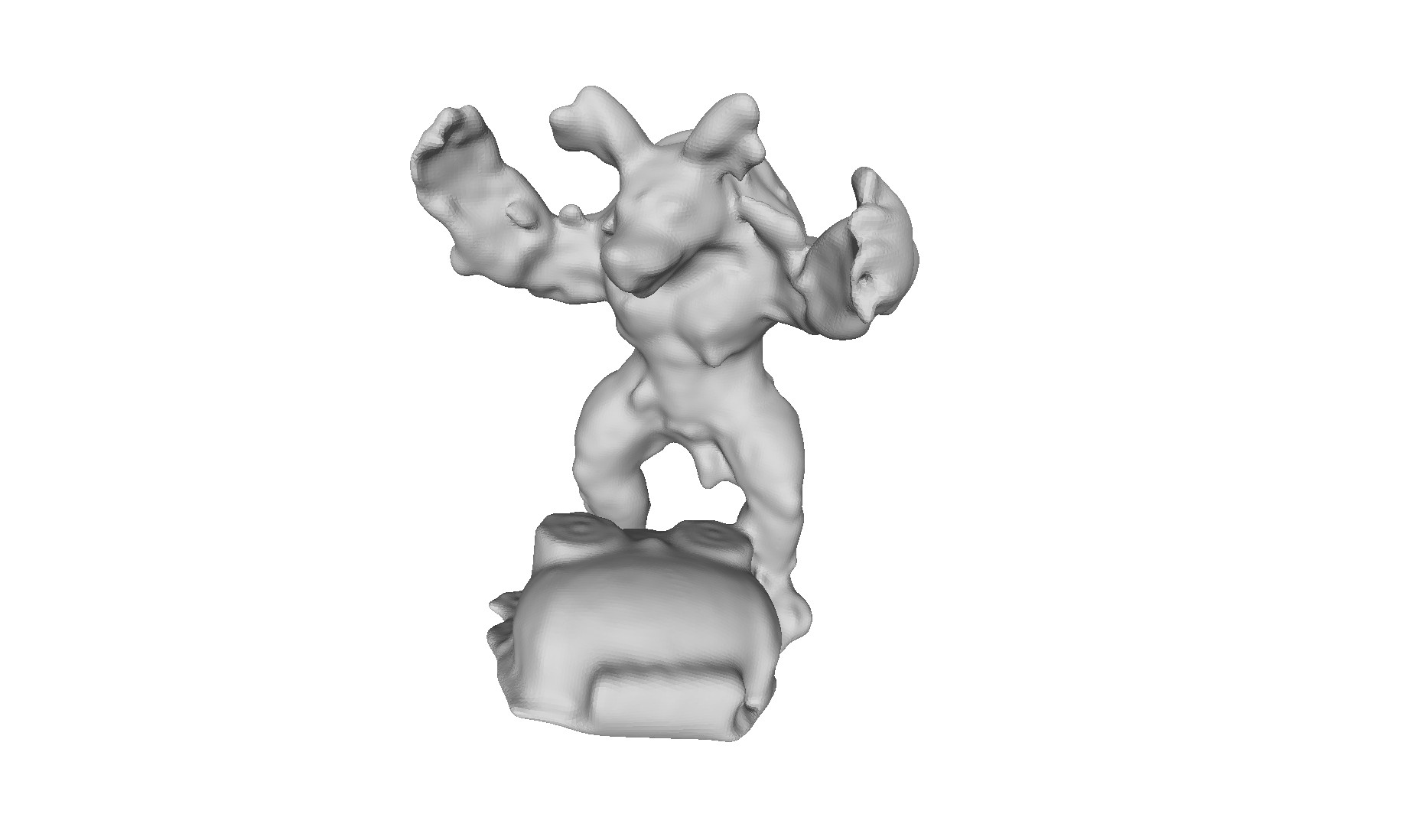}
   \includegraphics[clip,width=0.32\textwidth,trim={14.0cm 4.5cm 17.5cm  2.0cm}]{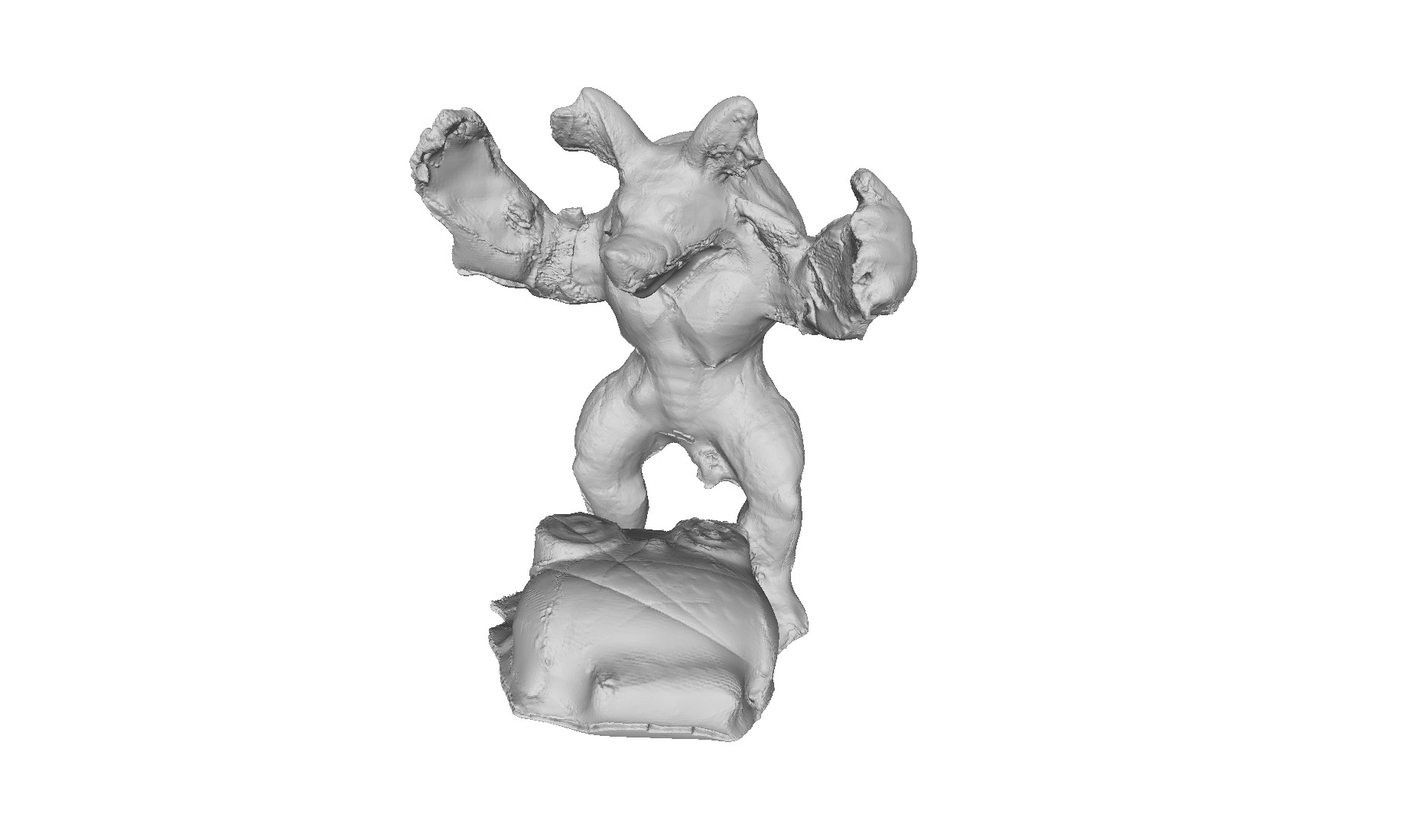}
   \includegraphics[clip,width=0.32\textwidth,trim={14.0cm 4.5cm 17.5cm  2.0cm}]{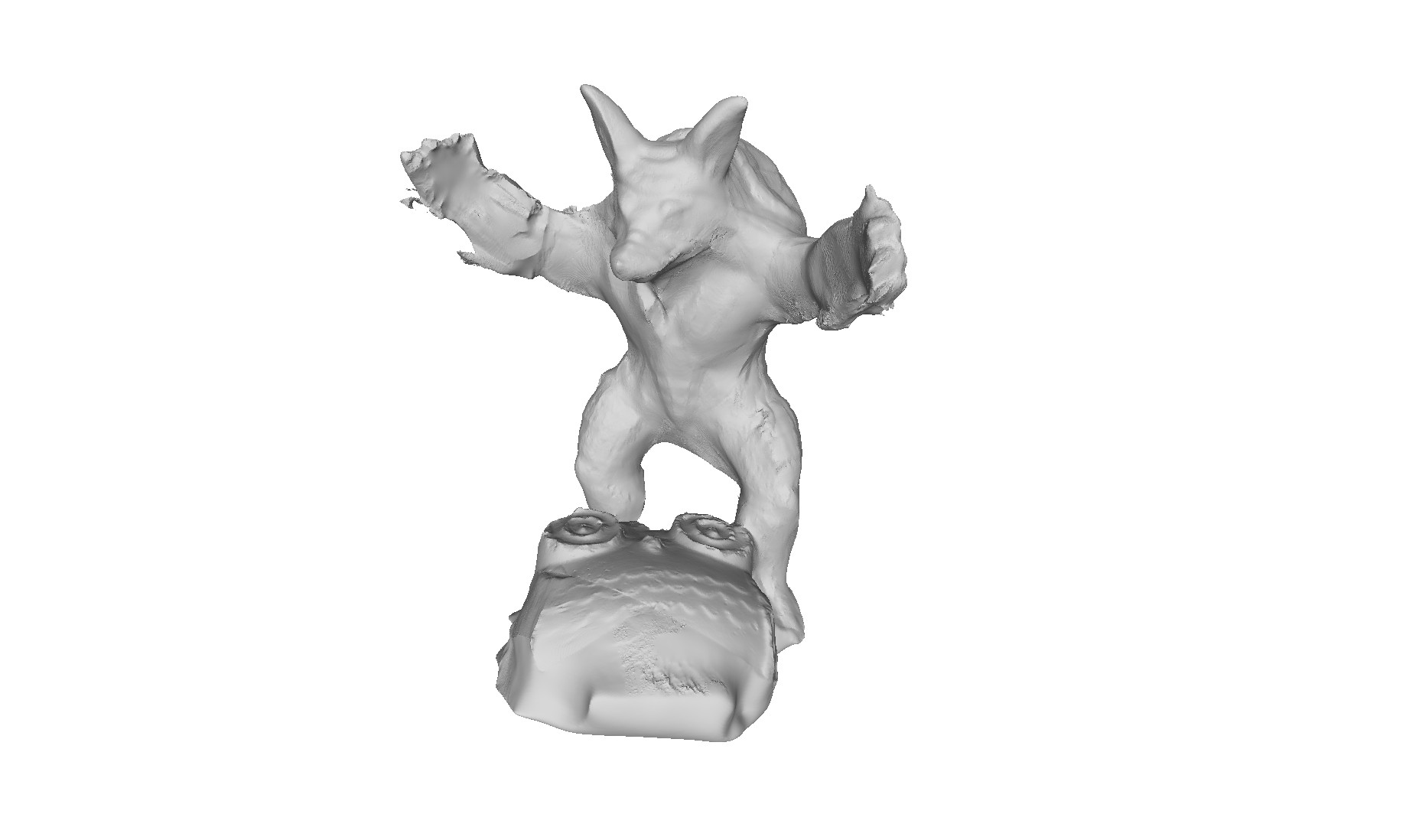}
 \caption{Top view of the reconstruction of the armadillo dataset. This includes the initial geometry estimate obtained with SFM(\cite{schoenberger2016sfm,schoenberger2016mvs} (\textit{left}),  \cite{Park2017} (\textit{middle}) and the proposed method  (\textit{right}). }
    \label{fig:owl_arma_top}
  \end{figure*}

\begin{figure}[b]
	\centering  
   \includegraphics[clip,width=0.28\textwidth,trim={8.25cm 4.5cm 12.5cm  2.0cm}]{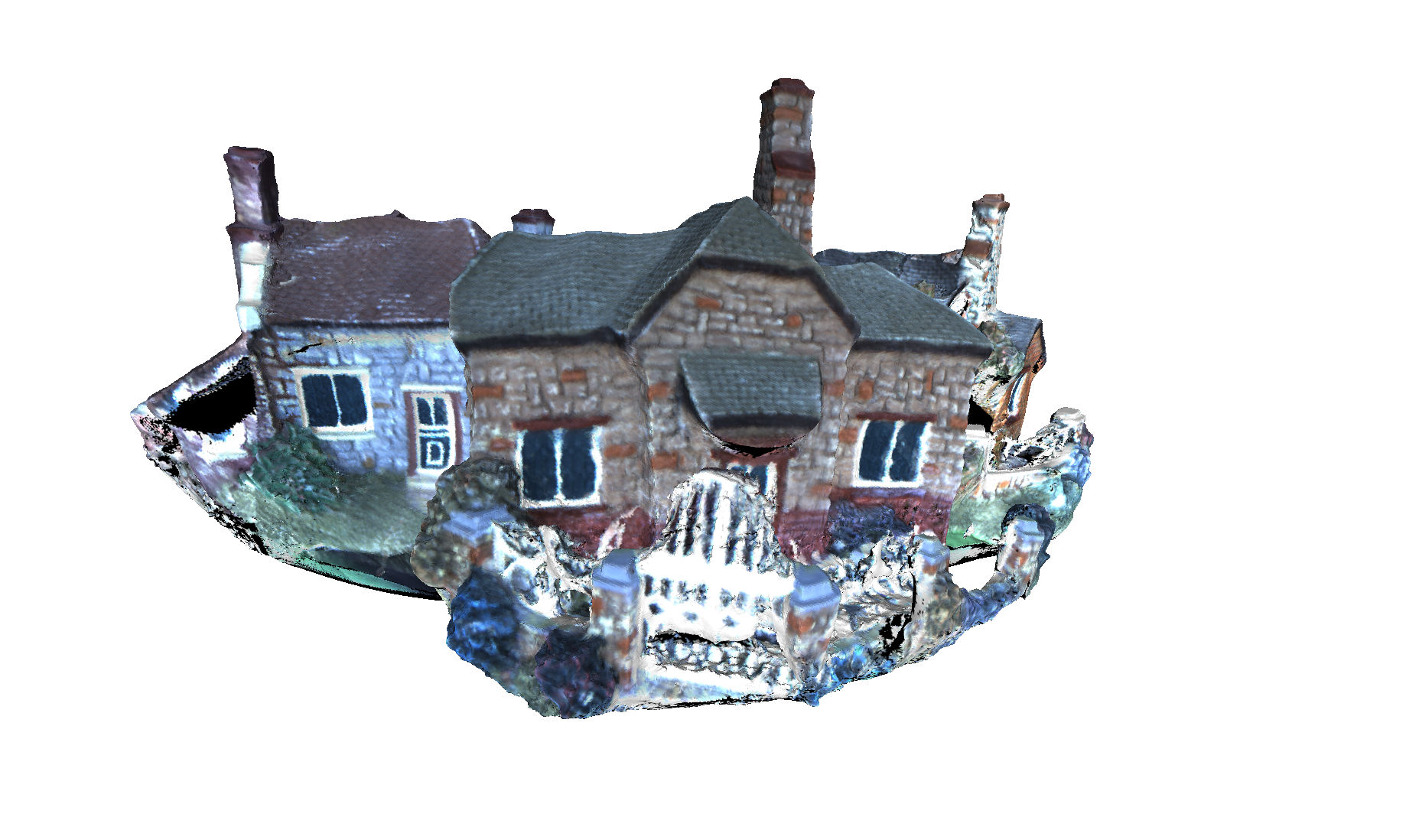}
      \includegraphics[clip,width=0.19\textwidth,trim={15cm 4cm 12cm  1.5cm}]{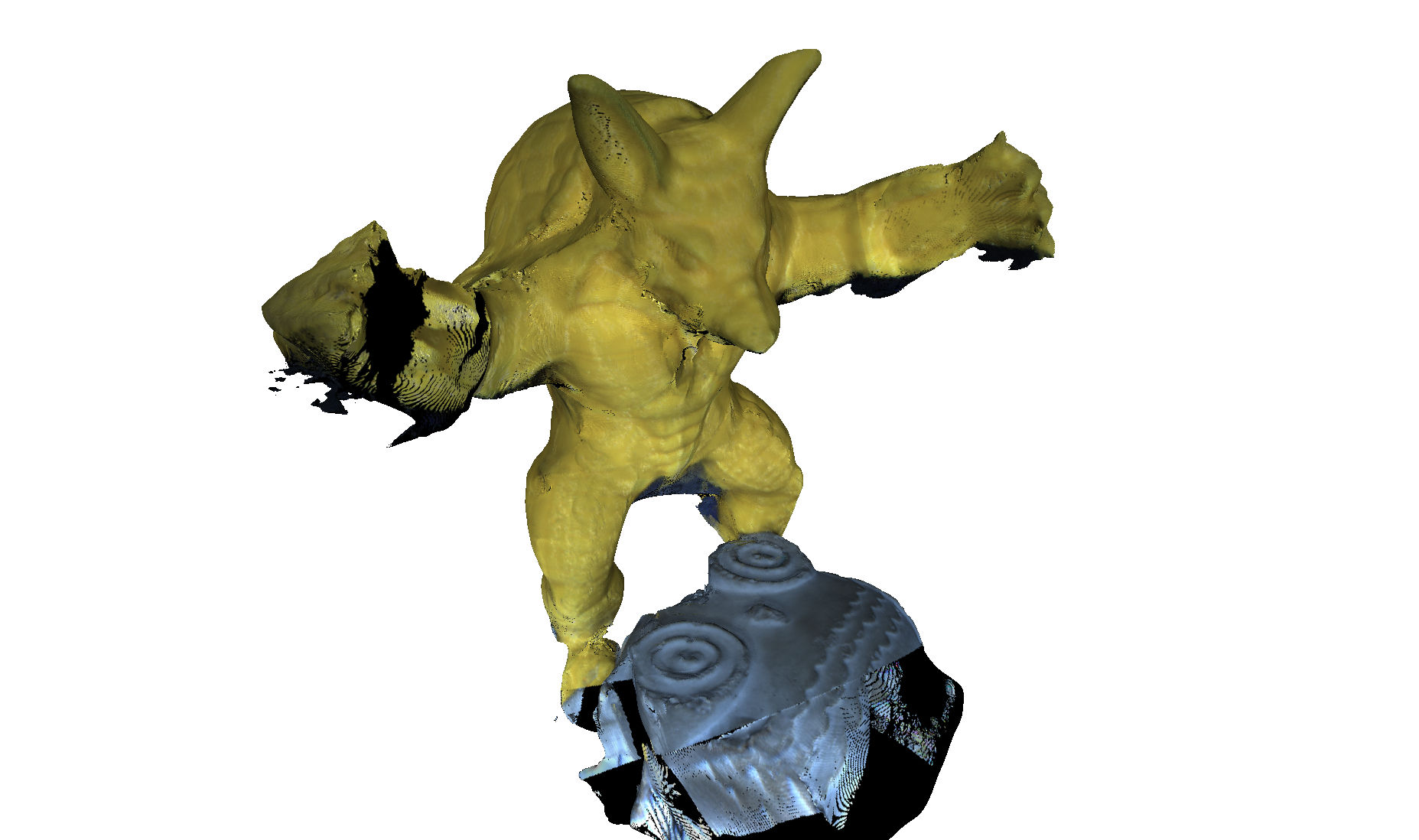}
 \caption{Albedo-mapped reconstruction of the 2 multi-object datasets. The albedo is calculated using simple least squares on Equation~(\ref{eq:irradi}) using the final geometry estimate (along with occlusion/shadow maps). Black albedo regions signify that this particular part of the surface is not visible in any  image.}
    \label{fig:recs_col} 
  \end{figure}

In addition, our methods performs very well on a challenging datasets where the existence of cast shadows highly degrades the quality of the reconstructions generated by \cite{Park2017} (see Figures~\ref{fig:house_comp}, \ref{fig:house_top}, \ref{fig:owl_arma_comp} and \ref{fig:owl_arma_top}). This is expected as our method ray-traces cast shadows whereas \cite{Park2017} only indirectly handles them using a robust estimation. Finally,  albedo colourised reconstructions can be seen in Figure~\ref{fig:recs_col}.

\section{Conclusion}
\label{sec:conclusion}

We presented the first volumetric parameterisation based on the signed distance function for the MVPS problem. Very high accuracy is achieved by using an octree implementation for processing and ray-tracing the volume on a tree. While considering photometric stereo images, our fully differential formulation is albedo independent as it uses the irradiance equation ratio approach for the near-field photometric stereo presented in \cite{Mecca2014near}. 

The main limitation of the proposed approach is the inability to cope with missing big portions of scene (this also true for most competing approaches e.g. \cite{Park2017,Zhou2013multi,Wu2014shading}). This is because Equation~(\ref{eq:normal_para}) only applies in the vicinity of a surface so the geometry cannot move very far away from initial estimate. In addition, the initial estimate is used for calculating the initial occlusion/shadow maps hence if an object that casts a shadow on the scene is not included in the initial estimate, the reconstruction of the shaded part will be problematic.


The main drawback of our method compared to mesh parameterisation techniques (e.g. \cite{Park2017}) is the elevated memory requirements. 
Even though the octree implementation minimises the number of voxels required, it is inevitable to need a few voxels per each potential surface point. This is due to the fact that the surface is the zero crossings of the SDF and at least a pair of opposite signed values are required per surface point. In addition, the use of the variational optimisation is also memory expensive as the matrix enconding the neighbouring information about voxels ($G$ in Equation~(\ref{eq:optim_full})) needs to be stored in memory as well.

As future work, the image ratio based modeling can be extended in order to handle specular highlights using the model presented in \cite{MeccaQLC2016}. This requires to enhance the variational solver with the inclusion of a shininess parameter, as an additional unknown per voxel. Additional realistic effects such as ambient light (\cite{logothetis2016near}) can also be included in the proposed model.



{\small
\bibliographystyle{ieee}
\bibliography{main}

\begin{thebibliography}{10}\itemsep=-1pt

\bibitem{Ackermann2014}
J.~Ackermann, F.~Langguth, S.~Fuhrmann, A.~Kuijper, and M.~Goesele.
\newblock Multi-view photometric stereo by example.
\newblock In {\em 2014 2nd International Conference on 3D Vision (3DV)},
  volume~1, pages 259--266, 2014.

\bibitem{Alldrin2007Towards}
N.~G. Alldrin and D.~J. Kriegman.
\newblock Toward reconstructing surfaces with arbitrary isotropic reflectance:
  A stratified photometric stereo approach.
\newblock In {\em 2007 IEEE International Conference on Computer Vision
  (ICCV)}, pages 1--8, 2007.

\bibitem{Barron2015}
J.~T. Barron and J.~Malik.
\newblock Shape, illumination, and reflectance from shading.
\newblock {\em {IEEE} Transactions on Pattern Analysis and Machine Intelligence
  ({PAMI})}, 37(8):1670--1687, 2015.

\bibitem{Basri2001b}
R.~Basri and D.~Jacobs.
\newblock {Photometric stereo with general, unknown lighting}.
\newblock In {\em IEEE Conference on Computer Vision and Pattern Recognition
  (CVPR)}, volume~2, pages 374--381, Kauai, USA, 2001.

\bibitem{Beljan2012}
M.~Beljan, J.~Ackermann, and M.~Goesele.
\newblock Consensus multi-view photometric stereo.
\newblock In {\em Joint DAGM (German Association for Pattern Recognition) and
  OAGM Symposium}, pages 287--296, 2012.

\bibitem{Blake1985}
A.~Blake, A.~Zisserman, and G.~Knowles.
\newblock Surface descriptions from stereo and shading.
\newblock {\em Image Vision Comput.}, 1985.

\bibitem{Delaunoy2011}
A.~Delaunoy and E.~Prados.
\newblock Gradient flows for optimizing triangular mesh-based surfaces:
  Applications to 3{D} reconstruction problems dealing with visibility.
\newblock {\em International Journal of Computer Vision {(IJCV)}},
  95(2):100--123, 2011.

\bibitem{GuoTOG2017}
K.~Guo, F.~Xu, T.~Yu, X.~Liu, Q.~Dai, and Y.~Liu.
\newblock Real-time geometry, albedo, and motion reconstruction using a single
  rgb-d camera.
\newblock {\em ACM Transactions on Graphics (ToG)}, 36(3), June 2017.

\bibitem{Hager1998}
G.~D. Hager and P.~N. Belhumeur.
\newblock Efficient region tracking with parametric models of geometry and
  illumination.
\newblock {\em IEEE Transactions on Pattern Analysis and Machine Intelligence
  (PAMI)}, 20(10):1025--1039, 1998.

\bibitem{Harltey2006}
A.~Harltey and A.~Zisserman.
\newblock {\em Multiple view geometry in computer vision {(2.} ed.)}.
\newblock Cambridge University Press, 2006.

\bibitem{Haussecker2001}
H.~W. Haussecker and D.~J. Fleet.
\newblock Computing optical flow with physical models of brightness variation.
\newblock {\em IEEE Transactions on Pattern Analysis and Machine Intelligence
  (PAMI)}, 23(6):661--673, 2001.

\bibitem{Hernandez2008Multiview}
C.~Hern{\'a}ndez, G.~Vogiatzis, and R.~Cipolla.
\newblock Multiview photometric stereo.
\newblock {\em IEEE Transactions on Pattern Analysis and Machine Intelligence
  {(PAMI)}}, 30(3):548--554, 2008.

\bibitem{Higo2009}
T.~Higo, Y.~Matsushita, N.~Joshi, and K.~Ikeuchi.
\newblock A hand-held photometric stereo camera for 3-{D} modeling.
\newblock In {\em 2009 IEEE International Conference on Computer Vision
  (ICCV)}, pages 1234--1241, 2009.

\bibitem{Horn1}
B.~K.~P. Horn.
\newblock Obtaining shape from shading information.
\newblock {\em The Psychology of Computer Vision, Winston, P. H. (Ed.)}, pages
  115--155, 1975.

\bibitem{Jin2001}
H.~Jin, P.~Favaro, and S.~Soatto.
\newblock Real-time feature tracking and outlier rejection with changes in
  illumination.
\newblock In {\em Proceedings Eighth IEEE International Conference on Computer
  Vision. ICCV}, 2001.

\bibitem{Jin2003}
H.~Jin, S.~Soatto, and A.~J. Yezzi.
\newblock Multi-view stereo beyond {L}ambert.
\newblock In {\em 2003 IEEE Conference on Computer Vision and Pattern
  Recognition ({CVPR})}, volume~1, pages I--I, 2003.

\bibitem{Joshi2007}
N.~Joshi and D.~J. Kriegman.
\newblock Shape from varying illumination and viewpoint.
\newblock In {\em 2007 IEEE International Conference on Computer Vision
  (ICCV)}, pages 1--7, 2007.

\bibitem{kazhdan2006poisson}
M.~Kazhdan, M.~Bolitho, and H.~Hoppe.
\newblock Poisson surface reconstruction.
\newblock In {\em Proceedings of the fourth Eurographics symposium on Geometry
  processing}, pages 61--70, 2006.

\bibitem{kazhdan2007unconstrained}
M.~Kazhdan, A.~Klein, K.~Dalal, and H.~Hoppe.
\newblock Unconstrained isosurface extraction on arbitrary octrees.
\newblock In {\em Eurographics Symposium on Geometry Processing}, 2007.

\bibitem{lambert1760}
J.~H. Lambert.
\newblock {\em Photometria sive De mensura et gradibus luminis, colorum et
  umbrae}.
\newblock Sumptibus viduae Eberhardi Klett, typis Christophori Petri
  Detleffsen, 1760.

\bibitem{Lim2005}
J.~Lim, J.~Ho, M.~Yang, and D.~J. Kriegman.
\newblock Passive photometric stereo from motion.
\newblock In {\em 2005 {IEEE} International Conference on Computer Vision
  (ICCV)}, pages 1635--1642, 2005.

\bibitem{Liu2008}
L.~Liu, L.~Zhang, Y.~Xu, C.~Gotsman, and S.~J. Gortler.
\newblock A local/global approach to mesh parameterization.
\newblock {\em Comput. Graph. Forum}, 2008.

\bibitem{logothetis2017semi}
F.~Logothetis, R.~Mecca, and R.~Cipolla.
\newblock Semi-calibrated near field photometric stereo.
\newblock In {\em 2017 IEEE Conference on Computer Vision and Pattern
  Recognition (CVPR)}, volume~3, page~8, 2017.

\bibitem{logothetis2016near}
F.~Logothetis, R.~Mecca, Y.~Qu{\'e}au, and R.~Cipolla.
\newblock Near-field photometric stereo in ambient light.
\newblock In {\em British Machine Vision Conference (BMVC)}, 2016.

\bibitem{luebke2003level}
D.~Luebke, M.~Reddy, J.~D. Cohen, A.~Varshney, B.~Watson, and R.~Huebner.
\newblock {\em Level of Detail for 3D Graphics}.
\newblock Morgan Kaufmann, 2003.

\bibitem{Maier2017}
R.~Maier, K.~Kim, D.~Cremers, J.~Kautz, and M.~Niessner.
\newblock {Intrinsic3D}: High-quality {3D} {R}econstruction by joint appearance
  and geometry optimization with spatially-varying lighting.
\newblock In {\em 2017 IEEE International Conference on Computer Vision
  (ICCV)}, 2017.

\bibitem{MalladiSV95}
R.~Malladi, J.~A. Sethian, and B.~C. Vemuri.
\newblock Shape modeling with front propagation: {A} level set approach.
\newblock {\em {IEEE} Trans. Pattern Anal. Mach. Intell.}, 17(2):158--175,
  1995.

\bibitem{MeccaQLC2016}
R.~Mecca, Y.~Qu\'eau, F.~Logothetis, and R.~Cipolla.
\newblock A single lobe photometric stereo approach for heterogeneous material.
\newblock {\em SIAM Journal on Imaging Sciences}, 9(4):1858--1888, 2016.

\bibitem{Mecca2014near}
R.~Mecca, A.~Wetzler, A.~Bruckstein, and R.~Kimmel.
\newblock {Near Field Photometric Stereo with Point Light Sources}.
\newblock {\em SIAM Journal on Imaging Sciences}, 7(4):2732--2770, 2014.

\bibitem{Moses2009}
Y.~Moses and I.~Shimshoni.
\newblock 3d shape recovery of smooth surfaces: Dropping the fixed-viewpoint
  assumption.
\newblock {\em IEEE Transactions on Pattern Analysis and Machine Intelligence
  (PAMI)}, 31(7):1310--1324, 2009.

\bibitem{nehab2005efficiently}
D.~Nehab, S.~Rusinkiewicz, J.~Davis, and R.~Ramamoorthi.
\newblock Efficiently combining positions and normals for precise 3d geometry.
\newblock {\em ACM transactions on graphics (TOG)}, 24(3):536--543, 2005.

\bibitem{NieBner2013}
M.~Nie{\ss}ner, M.~Zollh{\"o}fer, S.~Izadi, and M.~Stamminger.
\newblock Real-time 3d reconstruction at scale using voxel hashing.
\newblock {\em ACM Transactions on Graphics (ToG)}, 32(6):169, 2013.

\bibitem{Okatani2012}
T.~Okatani and K.~Deguchi.
\newblock Optimal integration of photometric and geometric surface measurements
  using inaccurate reflectance/illumination knowledge.
\newblock In {\em 2012 {IEEE} Conference on Computer Vision and Pattern
  Recognition (CVPR)}, pages 254--261, 2012.

\bibitem{osher2006level}
S.~Osher and R.~Fedkiw.
\newblock {\em Level Set Methods and Dynamic Implicit Surfaces}.
\newblock Applied Mathematical Sciences. Springer New York, 2006.

\bibitem{Park2017}
J.~Park, S.~N. Sinha, Y.~Matsushita, Y.~W. Tai, and I.~S. Kweon.
\newblock Robust multiview photometric stereo using planar mesh
  parameterization.
\newblock {\em {IEEE} Transactions on Pattern Analysis and Machine Intelligence
  {(PAMI)}}, 39(8):1591--1604, 2017.

\bibitem{queau2018led}
Y.~Qu{\'e}au, B.~Durix, T.~Wu, D.~Cremers, F.~Lauze, and J.-D. Durou.
\newblock Led-based photometric stereo: Modeling, calibration and numerical
  solution.
\newblock {\em Journal of Mathematical Imaging and Vision {(JMIV)}},
  60(3):313--340, 2018.

\bibitem{Sabzevari2012}
R.~Sabzevari, A.~D. Bue, and V.~Murino.
\newblock Multi-view photometric stereo using semi-isometric mappings.
\newblock In {\em 2012 Second International Conference on 3D Imaging, Modeling,
  Processing, Visualization Transmission}, 2012.

\bibitem{schoenberger2016sfm}
J.~L. Sch\"{o}nberger and J.-M. Frahm.
\newblock Structure-from-motion revisited.
\newblock In {\em Conference on Computer Vision and Pattern Recognition
  (CVPR)}, 2016.

\bibitem{schoenberger2016mvs}
J.~L. Sch\"{o}nberger, E.~Zheng, M.~Pollefeys, and J.-M. Frahm.
\newblock Pixelwise view selection for unstructured multi-view stereo.
\newblock In {\em European Conference on Computer Vision (ECCV)}, 2016.

\bibitem{seitz2006comparison}
S.~M. Seitz, B.~Curless, J.~Diebel, D.~Scharstein, and R.~Szeliski.
\newblock A comparison and evaluation of multi-view stereo reconstruction
  algorithms.
\newblock In {\em 2006 IEEE Conference on Computer Vision and Pattern
  Recognition (CVPR)}, pages 519--528, 2006.

\bibitem{Sheffer2006}
A.~Sheffer, E.~Praun, and K.~Rose.
\newblock Mesh parameterization methods and their applications.
\newblock {\em Foundations and Trends in Computer Graphics and Vision}, 2006.

\bibitem{Shi2018}
B.~Shi, Z.~Mo, Z.~Wu, D.~Duan, S.~K. Yeung, and P.~Tan.
\newblock A benchmark dataset and evaluation for non-lambertian and
  uncalibrated photometric stereo.
\newblock {\em IEEE Transactions on Pattern Analysis and Machine Intelligence},
  2018.

\bibitem{Treuille2004}
A.~Treuille, A.~Hertzmann, and S.~M. Seitz.
\newblock Example-based stereo with general {BRDF}s.
\newblock In {\em 8th European Conference on Computer Vision (ECCV)}, pages
  457--469, 2004.

\bibitem{Vlasic2009}
D.~Vlasic, P.~Peers, I.~Baran, P.~E. Debevec, J.~Popovic, S.~Rusinkiewicz, and
  W.~Matusik.
\newblock Dynamic shape capture using multi-view photometric stereo.
\newblock {\em {ACM} Transactions on Graphics}, 28(5), 2009.

\bibitem{Ward92}
G.~J. Ward.
\newblock Measuring and modeling anisotropic reflection.
\newblock {\em ACM SIGGRAPH Computer Graphics}, 26(2):265--272, 1992.

\bibitem{Weber2002}
M.~Weber, A.~Blake, and R.~Cipolla.
\newblock Towards a complete dense geometric and photometric reconstruction
  under varying pose and illumination.
\newblock In {\em British Machine Vision Conference (BMVC)}, 2002.

\bibitem{Woodham1980}
R.~J. Woodham.
\newblock Photometric method for determining surface orientation from multiple
  images.
\newblock {\em Optical Engineering}, 19(1):134--144, 1980.

\bibitem{WuLiu2011}
C.~Wu, Y.~Liu, Q.~Dai, and B.~Wilburn.
\newblock Fusing multiview and photometric stereo for 3{D} reconstruction under
  uncalibrated illumination.
\newblock {\em IEEE transactions on visualization and computer graphics},
  17(8):1082--1095, 2011.

\bibitem{Wu2011Shading}
C.~Wu, K.~Varanasi, Y.~Liu, H.~Seidel, and C.~Theobalt.
\newblock Shading-based dynamic shape refinement from multi-view video under
  general illumination.
\newblock In {\em 2011 IEEE International Conference on Computer Vision
  (ICCV)}, pages 1108--1115, 2011.

\bibitem{Wu2014shading}
C.~Wu, M.~Zollh\"{o}fer, M.~Nie\ss~ner, M.~Stamminger, S.~Izadi, and
  C.~Theobalt.
\newblock Real-time shading-based refinement for consumer depth cameras.
\newblock {\em ACM Transactions on Graphics (TOG)}, 33(6):200, 2014.

\bibitem{Zhang2003}
L.~Zhang, B.~Curless, A.~Hertzmann, and S.~M. Seitz.
\newblock Shape and motion under varying illumination: unifying structure from
  motion, photometric stereo, and multiview stereo.
\newblock In {\em 2003 IEEE International Conference on Computer Vision
  (ICCV)}, pages 618--625, 2003.

\bibitem{Zhou2013multi}
Z.~Zhou, Z.~Wu, and P.~Tan.
\newblock Multi-view photometric stereo with spatially varying isotropic
  materials.
\newblock In {\em 2013 IEEE Conference on Computer Vision and Pattern
  Recognition{(CVPR)}}, pages 1482--1489, 2013.

\bibitem{Zollhofer15}
M.~Zollh{\"o}fer, A.~Dai, M.~Innmann, C.~Wu, M.~Stamminger, C.~Theobalt, and
  M.~Nie{\ss}ner.
\newblock Shading-based refinement on volumetric signed distance functions.
\newblock {\em ACM Transactions on Graphics (TOG)}, 34(4):96, 2015.

\end{thebibliography}
}

\end{document}